\documentclass[default,iicol]{sn-jnl}

\usepackage{epsfig}
\usepackage{graphicx}
\usepackage{amsmath}
\usepackage{amssymb}
\usepackage{booktabs}
\usepackage{comment}
\usepackage{color}

\usepackage{latexsym}
\usepackage{makecell,diagbox}
\usepackage{algorithm}
\usepackage{listings}

\usepackage{amsfonts}
\usepackage{colortbl}
\usepackage{microtype}
\usepackage{multirow}
\usepackage{adjustbox}
\usepackage{enumitem}
\usepackage{float}
\usepackage{bm}
\usepackage{bbm} 

\usepackage{bbding}
\usepackage{pifont}
\usepackage{wasysym}
\usepackage{xcolor}

\jyear{2022}%

\theoremstyle{thmstyleone}%
\theoremstyle{thmstyletwo}%

\theoremstyle{thmstylethree}%

\begin{document}

\title[CLIP-Adapter]{CLIP-Adapter: Better Vision-Language Models with Feature Adapters}


\author[1]{\fnm{Peng} \sur{Gao}}\email{gaopeng@pjlab.org.cn}
\equalcont{These authors contributed equally to this work.}

\author[3]{\fnm{Shijie} \sur{Geng}}\email{sg1309@rutgers.edu}
\equalcont{These authors contributed equally to this work.}

\author[1,2]{\fnm{Renrui} \sur{Zhang}}\email{zhangrenrui@pjlab.org.cn}
\equalcont{These authors contributed equally to this work.}

\author[1]{\fnm{Teli} \sur{Ma}}

\author[2]{\fnm{Rongyao} \sur{Fang}}

\author[3]{\fnm{Yongfeng} \sur{Zhang}}

\author*[2,4]{\fnm{Hongsheng} \sur{Li}}\email{hsli@ee.cuhk.edu.hk}

\author*[1]{\fnm{Yu} \sur{Qiao}}\email{qiaoyu@pjlab.org.cn}

\affil[1]{\orgname{Shanghai AI Laboratory}, \city{Shanghai}, \country{China}}

\affil[2]{\orgname{CUHK MMLab}, \city{Hong Kong SAR}, \country{China}}

\affil[3]{\orgname{Rutgers University}, \state{New Jersey}, \country{US}}

\affil[4]{\orgname{Centre for Perceptual and Interactive Intelligence (CPII)}, \city{Hong Kong SAR}, \country{China}}


\abstract{Large-scale contrastive vision-language pretraining has shown significant progress in visual representation learning. Unlike traditional visual systems trained by a fixed set of discrete labels, a new paradigm was introduced in \cite{radford2021learning} to directly learn to align images with raw texts in an open-vocabulary setting. On downstream tasks, a carefully chosen text prompt is employed to make zero-shot predictions.~To avoid non-trivial prompt engineering, context optimization \cite{zhou2021coop} has been proposed to learn continuous vectors as task-specific prompts with few-shot training examples.~In this paper, we show that there is an alternative path to achieve better vision-language models other than prompt tuning.~While prompt tuning is for the textual inputs, we propose CLIP-Adapter to conduct fine-tuning with feature adapters on either visual or language branch. Specifically, CLIP-Adapter adopts an additional bottleneck layer to learn new features and performs residual-style feature blending with the original pretrained features.~As a consequence, CLIP-Adapter is able to outperform context optimization while maintaining a simple design. Experiments and extensive ablation studies on various visual classification tasks demonstrate the effectiveness of our approach. Code is available at \url{https://github.com/gaopengcuhk/CLIP-Adapter}.}

\keywords{Feature Adapter, Vision-Language Model, Few-shot learning, Open-Vocabulary}



\maketitle

\section{Introduction}

\begin{figure*}[t!]
\centering
\includegraphics[width=0.93\linewidth,trim={0cm 0cm 0cm 0cm}]{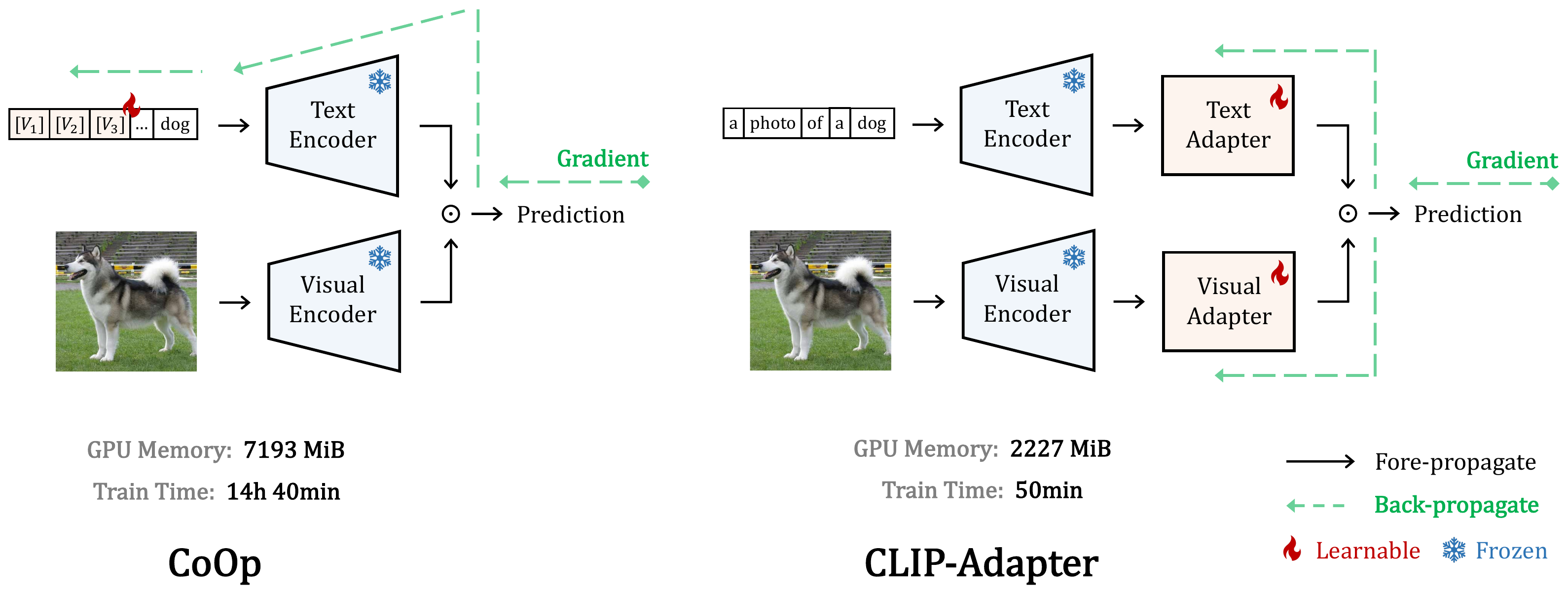}
\caption{Comparison of CoOp and our CLIP-Adapter. Instead of using learnable prompts, CLIP-Adapter adopts lightweight residual-style adapters after CLIP's text and visual encoders.
During training, CLIP-Adapter does not require to calculate and propagate the gradients through CLIP's encoder, which thus consumes less computational cost.}
\label{fig:back}
\end{figure*}

Visual understanding tasks, such as classification~\cite{krizhevsky2012imagenet,he2016deep,howard2017mobilenets,dosovitskiy2021vit,touvron2021training,gao2021container,mao2021dual}, object detection~\cite{ren2015faster,carion2020end,gao2021fast}, and semantic  segmentation~\cite{long2015fully}, have been improved significantly based on the better architecture designs and large-scale high-quality datasets. Unfortunately, collecting large-scale high-quality datasets for every visual task is labor-intensive and too expensive to scale.~To solve the problem, the ``pretraining-finetuning'' paradigm, namely pretraining on large-scale datasets like ImageNet~\cite{krizhevsky2012imagenet} and then fine-tuning on a variety of downstream tasks, has been widely adopted in vision domain.~However, such approaches still need a huge amount of annotations for fine-tuning on many downstream tasks. Recently, Contrastive Language-Image Pretraining (CLIP)~\cite{radford2021learning} was proposed for solving vision tasks by exploiting contrastive learning with large-scale noisy image-text pairs. It achieves inspirational performances on various visual classification tasks without any annotations (i.e., zero-shot transfer) by putting visual categories into suitable hand-crafted template as prompts.

Although prompt-based zero-shot transfer learning showed promising performances, designing good prompts remains an engineering problem that demands substantial time and domain knowledge.~To address the issue, Context Optimization (CoOp)~\cite{zhou2021coop} further proposed to learn continuous soft prompts with few-shot examples for replacing the carefully-chosen hard prompts. CoOp brings about significant improvement on few-shot classification over both zero-shot CLIP and linear probe CLIP settings, exhibiting the potential of prompt tuning on large-scale pretrained vision-language models. 
In this paper, we propose a different approach for better adapting vision-language models with feature adapters instead of prompt tuning.~Different from CoOp that performs soft prompt optimization, we simply conduct fine-tuning on the light-weight additional feature adapters. Because of the over-parameterization of CLIP and lack of enough training examples, naive finetuning would lead to overfitting on specific datasets and the training process would be very slow owing to the forward and backward propagations across all CLIP layers. 

\begin{figure*}[t!]
\centering
\includegraphics[height=0.72\linewidth,trim={0cm 0cm 0cm 0cm}]{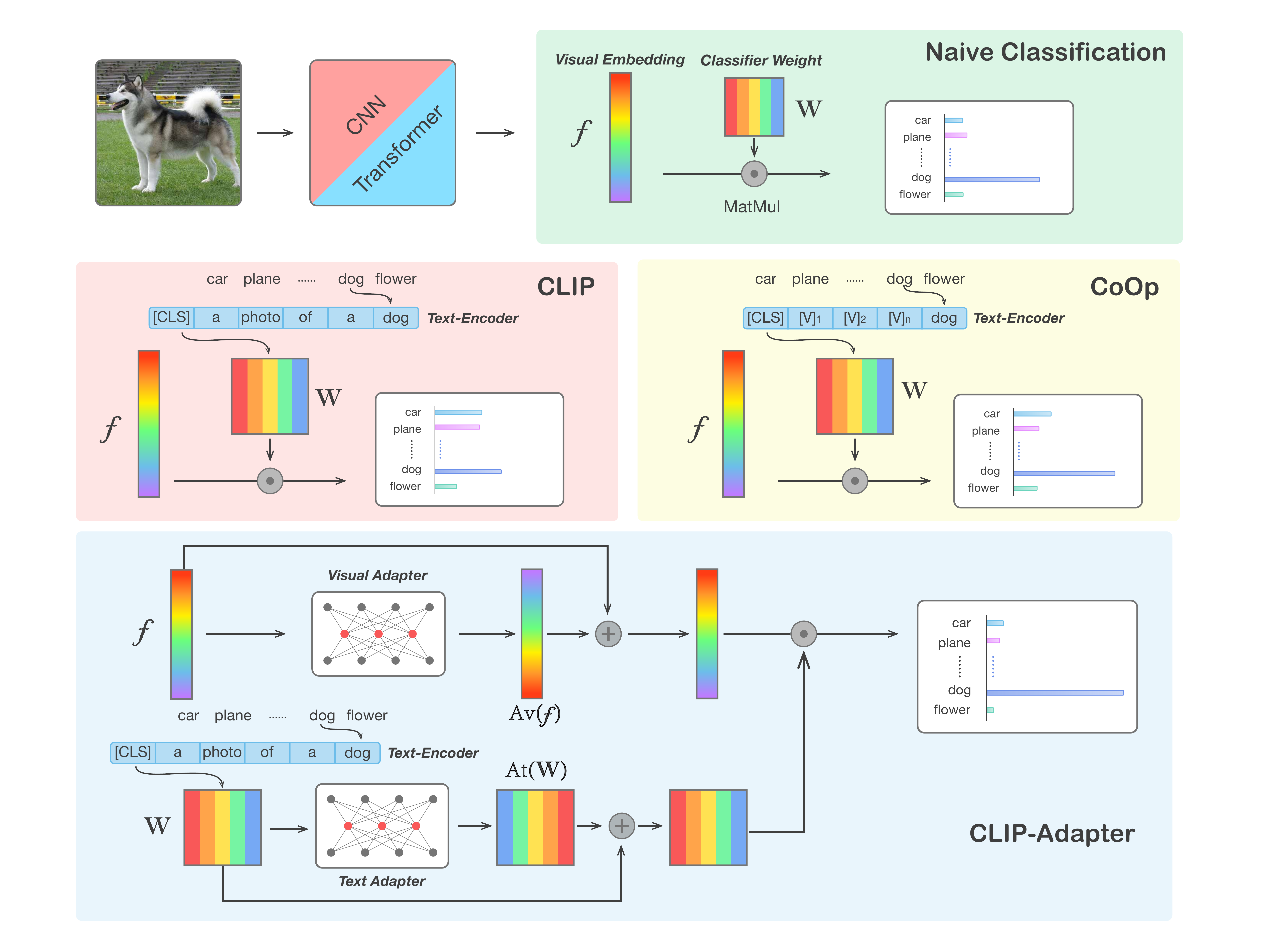}
\caption{Comparison of different visual classification architectures.~The image in the top row with a green region shows the naive pipeline for image classification~\cite{krizhevsky2012imagenet}, where $f$ and $\mathbf{W}$ represents the visual feature and classifier weight respectively. The following pink, yellow and blue regions represent the pipeline of CLIP~\cite{radford2021learning}, CoOp~\cite{zhou2021coop}, and our proposed CLIP-Adapter respectively.}
\label{fig:model}
\end{figure*}

Motivated by the adapter modules in parameter-efficient transfer learning~\cite{houlsby2019parameter}, we propose \emph{CLIP-Adapter}, which only finetunes a small number of additional weights instead of optimizing all parameters of CLIP, as shown in Figure~\ref{fig:back}. CLIP-Adapter adopts a lightweight bottleneck architecture to prevent the potential overfitting problem of few-shot learning by reducing the number of parameters. Meanwhile, CLIP-Adapter is different from~\cite{houlsby2019parameter} in two important aspects:
CLIP-Adapter only adds two additional linear layers following the last layer of vision or language backbone. This is because the original pretrained encoder of CLIP has already been equipped with strong representation capabilities, so it only requires a lightweight adaption in the form of residuals. In contrast, the original adapter modules are inserted into all layers of the language backbone; In addition, CLIP-Adapter mixes the original zero-shot visual or language embedding with the corresponding finetuning feature via residual connection. Through such a ``residual-style blending'', CLIP-Adapter can simultaneously exploit the knowledge stored in the original CLIP and the freshly learned knowledge originated from the few-shot training examples. Figure \ref{fig:model} gives an intuitive illustration of the differences between our CLIP-Adapter and other visual classification architectures. Overall, our contributions can be summarized as follows:
\begin{itemize}[leftmargin=*,noitemsep]
    \item We propose CLIP-Adapter that conducts residual-style feature blending to achieve efficient few-shot transfer learning via fine-tuning. 
    \item Compared with CoOp, CLIP-Adapter achieves better few-shot classification performance while having a much simpler design, demonstrating that CLIP-Adapter is a promising alternative to prompt tuning.
    \item We perform extensive ablation studies of CLIP-Adapter on eleven classification datasets to analyze its characteristics.
\end{itemize}

\section{Related Work}
\noindent\textbf{Model Fine-Tuning.~} Deep neural network is data-hungry.~However, collecting and annotating large amount of high-quality data is costly and even impossible for some special domains.~The ``pretraining-finetuning paradigm'' offers a good solution to different computer vision~\cite{krizhevsky2012imagenet,simonyan2015very,he2016deep} and natural language processing~\cite{devlin2019bert,dong2019unified,conneau2020unsupervised} tasks and has been widely adopted for many years.~For data-efficient finetuning over downstream tasks, adapter modules~\cite{houlsby2019parameter} is proposed to freeze the weight of backbones and insert learnable linear layers to each Transformer layer. Subsequent work such as Parallel Adapter \cite{he2022towards} and VL-Adapter \cite{sung2022vl} further extend \cite{houlsby2019parameter}'s ability to language and multimodal tasks. 
Some other methods, e.g., black-box tuning~\cite{sun2022black}, ladder side-tuning~\cite{sung2022lst}, sparse structure search\cite{hu2022sparse}, and Scaling\&Shifting~\cite{lian2022scaling} also introduce different techniques for adapting large language and vision models in a parameter-efficient manner.
Different from existing approaches, the proposed CLIP-Adapter applies a simple residual transformation layer over the feature embedding or classifier weight generated by CLIP. Thanks to the residual connection and bottleneck linear layer, CLIP-Adapter can improve the performance of CLIP on few-shot learning setting and achieve superior performance than the recently proposed CoOp.~To alleviate the performance gap under distribution shifting, WiSE-FT~\cite{wortsman2022robust} proposes a post-ensemble method for improving CLIP's out-of-distribution robustness.~While our CLIP-Adapter adopts a learnable gating ratio to dynamically balance and mix the knowledge from the original features and CLIP-Adapter's outputs throughout the training stage.

\vspace{3pt}
\noindent\textbf{Prompt Design.~} Prompt design~\cite{liu2023pre} are popularized by the success of GPT series~\cite{radford2019language,brown2020language}. GPT-3 showed that a huge autoregressive language model trained on a large-scale dataset can perform any NLP tasks in a zero-shot or few-shot style without finetuning the base architecture.~Following the brand new ``pretrain, prompt, and predict'' paradigm~\cite{liu2023pre}, various prompt design approaches are proposed recently. As the earliest attempt, one type of them focus on tuning pretrained language or vision-language models with natural language discrete prompts~\cite{tsimpoukelli2021multimodal,alayrac2022flamingo,wang2022simvlm,yao-etal-2022-pevl,yao2021cpt,gao2021making} or prompt engineering by mining or generating proper natural language discrete prompts~\cite{jiang2020can,shin2020autoprompt,gao2021making}. In contrast, continuous prompts circumvent the restriction from pretrained language models and are adopted by various approaches such as~\cite{gu2022ppt,li2021prefix,liu2021gpt,lester2021power} on NLP tasks.~Recently, they have also been introduced to vision tasks~\cite{jia2022visual}. Motivated by GPT-3, CLIP trains a large contrastive learning model over 400 million image-text pairs and demonstrates the potential for prompt-based zero-shot visual classification. With CLIP as the backbone, CoOp~\cite{zhou2021coop} further shows that optimizing continuous prompts can largely surpass manually-designed discrete prompts on vision tasks. In this paper, we demonstrate that prompt tuning is not the only path to better vision-language models. Fine-tuning with a small portion of parameters can also achieve comparable or even better performance on vision tasks yet with much simpler design.

\vspace{3pt}
\noindent\textbf{Vision-Language Models.~} Exploring the interaction between vision and language is a core research topic in artificial intelligence.~Previously, attention-based approaches such as bottom-up top-down attention~\cite{anderson2018bottom}, BAN~\cite{kim2018bilinear}, Intra-Inter~\cite{gao2019dynamic}, and MCAN~\cite{yu2019deep} had dominated visual-language tasks. Inspired by the success of BERT~\cite{devlin2019bert}, ViLBERT~\cite{lu2019vilbert}, LXMERT~\cite{tan2019lxmert}, UNITER~\cite{chen2020uniter}, Oscar~\cite{li2020oscar}, ALBEF~\cite{li2021align}, and BEiT~\cite{wang2022image} further push the boundary of multimodal reasoning. Recently, CLIP~\cite{radford2021learning} and ALIGN~\cite{jia2021scaling} demonstrates the power of visual-language contrastive representation learning. They achieve astonishing results on a wide spectrum of vision tasks, including 3D~\cite{zhu2022pointclip,zhang2023learning}, video~\cite{lin2022frozen}, and depth~\cite{zhang2022can} understanding. To further close the gap between CLIP and supervised training, CoOp proposes a continuous prompt optimization method for improving the performance on visual classification tasks. While CoOp improves vision-language models from the perspective of prompt design, our CLIP-Adapter explores simple finetuning with lightweight feature adapters.

\section{Our Approach}
In this section, we introduce the proposed CLIP-Adapter. In Section~\ref{sec:view}, we first revisit CLIP and CoOp from the perspective of classifier weight generation.~In Section~\ref{sec:clip-adapter}, we elaborate the details of the proposed CLIP-Adapter.~In Section~\ref{sec:variants}, we provide several variants of CLIP-Adapter.

\subsection{Classifier Weight Generation for Few-Shot Learning}
\label{sec:view}
Let us first review the basic framework for image classification using deep neural networks: Given an image $\mathbf{I} \in \mathbb{R}^{H \times W \times 3}$, where $H$ and $W$ stands for the height and width of the image respectively, a neural network backbone that consists of cascade of basic components (e.g., CNN, Transformer~\cite{vaswani2017attention} or the mixture of both) takes $\mathbf{I}$ and transforms it into a feature manifold $f \in \mathbb{R}^{D}$, where $D$ represents the feature dimensionality.~To perform classification, the image feature vector $f$ is then multiplied with a classifier weight matrix $\mathbf{W} \in \mathbb{R}^{D \times K}$, where $K$ represents the number of classes to be classified.~After matrix multiplication, we can obtain a $K$-dimensional logit. A Softmax function is used to convert the logit into a probability vector $p \in \mathbb{R}^{K}$ over the $K$ classes. The whole process can be written as the following equations:
\begin{equation}
        f = \mathrm{Backbone}(\mathbf{I}),~~p_i = \frac
        {\operatorname{exp}(\mathbf{W}_{i}^T f) / \tau}
        {\sum_{j=1}^{K} \operatorname{exp}(\mathbf{W}_{j}^T f) / \tau}, \label{probability} 
\end{equation}
where $\tau$ stands for the temperature of Softmax, $\mathbf{W}_i$ represents the prototype weight vector for class $i$, and $p_i$ denotes the probability of category $i$.

Different from supervised training, in the paper, we are interested in image classification with few-shot examples. Training the backbone and classifier together from scratch with a small number of samples is prone to overfit certain datasets and might suffer from severe performance drop on the test split. Typically, the representative paradigm on few-shot learning is to first pretrain the backbone on a large-scale dataset, and then transfer the learned knowledge to downstream tasks by either conducting zero-shot prediction directly or further fine-tuning on few-shot examples. 

CLIP adheres to the zero-shot transfer style -- it first pretrains the visual backbone and textual encoder through contrastive learning on large-scale noisy image-text pairs, and then after pretraining, CLIP directly performs image classification without any finetuning.~Given an image classification downstream dataset that contains $K$ categories with their natural language name $\{C_{1}, \dots, C_k\}$, CLIP constructs to place each category name $C_i$ into the pre-defined hard prompt template $H$. Then the language feature extractor encodes the resulting prompt as a classifier weight $\mathbf{W}_i$. We denote the classifier weight generation process as below:
\begin{equation}
        \mathbf{W}_{i} = \operatorname{Text-Encoder}(\mathrm{Tokenizer}([H; C_{i}])).
        \label{eq:weight_clip}
\end{equation}

Alternatively, CoOp adopts continuous prompts instead of hand-crafted hard prompts. CoOp creates a list of random-initialized learnable soft tokens $S \in \mathbb{R}^{L \times D}$, where $L$ stands for the length of the soft token sequence. The soft token sequence $S$ is then concatenated to each class name $C_i$ and thus form a prompt. We represent the whole process as 
\begin{equation}
    \label{eq:weight_coop}
    \begin{aligned}
        \mathbf{W}_{i} &= \operatorname{Text-Encoder}([S; \mathrm{Tokenizer}(C_{i})]).
    \end{aligned}
\end{equation}

For both CLIP and CoOp, with the generated classifier weight $\mathbf{W}_i$, where $i \in \{1, \cdots, K\}$, we can thus calculate the prediction probability $p_i$ for class $i$ by the previously mentioned Eq.~\eqref{probability}.

\subsection{CLIP-Adapter}
\label{sec:clip-adapter}
Unlike CoOp's prompt tuning, we present an alternative framework for achieving better vision-language models on few-shot image classification by fine-tuning additional feature adapters.~We claim that the previous widely-adopted ``pretrain-finetuning'' paradigm would fail in finetuning the whole CLIP backbone under the few-shot setting due to the enormous amount of parameters and the shortage of training examples. Hence, we propose CLIP-Adapter, which only appends a small number of additional learnable bottleneck linear layers to CLIP's language and image branches while keeping the original CLIP backbone frozen during few-shot fine-tuning. However, naive fine-tuning with additional layer may still fall into overfitting on the few-shot examples. To deal with overfitting and improve the robustness of CLIP-Adapter, we further adopt residual connections to dynamically blend the fine-tuned knowledge with the original knowledge from CLIP's backbone. 

Specifically, given the input image $\mathbf{I}$ and a set of categories' natural language names $\{C_i\}^{K}_{i=1}$, the image feature $f$ and classifier weight $\mathbf{W}$ from the original CLIP backbone are computed with Equations~\eqref{probability} and~\eqref{eq:weight_clip}. 
Afterwards, two learnable feature adapters, $A_v(\cdot)$ and $A_t(\cdot)$, each of which contains two layers of linear transformations, are integrated to transform $f$ and $\mathbf{W}$, respectively. 
We adopt a residual connection for the feature adapter to avoid forgetting the original knowledge encoded by the pretrained CLIP. Two constant values $\alpha$ and $\beta$ are employed as ``residual ratio'' to help adjust the degree of maintaining the original knowledge for better performance. In summary, the feature adapters can be written as
\begin{align}
        A_v(f) &= \operatorname{ReLU}(f^{T} \mathbf{W}_1^v) \mathbf{W}_2^v, \\
        A_t(\mathbf{W}) &= \operatorname{ReLU}(\mathbf{W}^{T} \mathbf{W}_1^t) \mathbf{W}_2^t,
\end{align}
where $\mathbf{W}_1^v$, $\mathbf{W}_2^v$ and $\mathbf{W}_1^t$, $\mathbf{W}_2^t$ are the weights of bottleneck linear layers for visual branch and text branch, respectively.
The new knowledge captured via finetuning is added with the original features via residual connections:
\begin{align}
        f^{\star} &= \alpha A_{v}(f)^{T} + (1-\alpha) f, \\
        \mathbf{W}^{\star} &= \beta A_{t}(\mathbf{W})^{T} + (1 - \beta) \mathbf{W}.
\end{align}
After obtaining new image feature $f^{\star}$ and classifier weight $\mathbf{W}^{\star}$, we also adopt Equation~\eqref{probability} to calculate the category probability vector $P = \{p_i\}^{K}_{i=1}$ and predict the image category by selecting the class $\hat{i}$ that has the highest probability: $\hat{i} = \mathop{\arg\max}_{i} p_{i}$.

During the few-shot training, the weights of $A_v(\cdot)$ and $A_t(\cdot)$ are optimized with the contrastive loss following original CLIP~\cite{radford2021learning} as below:
\begin{equation}
\mathcal{L}_{\theta} = -\frac{1}{N} \sum_{i}^{N} \log \frac{\exp \left(\mathbf{W}_{i}^{\top} f_{i} / \tau\right)}{\sum_{j=1}^{N} \exp \left(\mathbf{W}_{j}^{\top} f_{i} / \tau\right)}, \nonumber
\end{equation}
where $N$ is the total number of training examples; $\theta = \{\mathbf{W}_1^v, \mathbf{W}_2^v, \mathbf{W}_1^t, \mathbf{W}_2^t\}$ represents all learnable parameters of CLIP-Adapter.

\subsection{Variants of CLIP-Adapter}
\label{sec:variants}

Our CLIP-Adapter has three structural variants: 1) only fine-tuning the feature adapter for the image branch while keeping the text branch frozen; 2) only fine-tuning the feature adapter for the text branch while keeping the image branch frozen; 3) fine-tuning both the image and text branches of CLIP backbone.~In terms of the hyperparameters $\alpha$ and $\beta$, we observe that different datasets have different optimal $\alpha$ and $\beta$ values. Choosing the hyperparameters manually is time-consuming and laborious. Thus we also explore learning $\alpha$ and $\beta$ in a differentiable manner by setting them as learnable parameters. In this way, $\alpha$ and $\beta$ can be dynamically predicted from either visual feature or classifier weight via a hypernetwork $Q$: $\alpha, \beta = Q(f, \mathbf{W})$.

\section{Experiments}
\subsection{Experimental Setups}

\noindent\textbf{Datasets.~}
Following CLIP and CoOp, we select 11 image classification datasets to validate CLIP-Adapter's effectiveness, namely ImageNet~\cite{deng2009imagenet}, StanfordCars~\cite{krause20133d}, UCF101 \cite{soomro2012ucf101}, Caltech101~\cite{fei2004learning}, Flowers102~\cite{nilsback2008automated}, SUN397~\cite{xiao2010sun}, DTD~\cite{cimpoi2014describing}, EuroSAT~\cite{helber2019eurosat}, FGVCAircraft~\cite{maji2013fine}, OxfordPets~\cite{parkhi2012cats}, and Food101~\cite{bossard2014food}. Specifically, we train our CLIP-Adapter under the few-shot setups of $1$, $2$, $4$, $8$, $16$ shots and then test the tuned models on full test splits. Considering the randomness of few-shot training, we run every setting three times and report the average accuracy. We conduct all experiments on a single NVIDIA A100 GPU.

\vspace{3pt}
\noindent\textbf{Implementation Details.~} \textit{The first variant of CLIP-Adapter is adopted by default if not specified, which finetunes the image feature while freezing the classifier weight. In other words, it only implements CLIP-Adapter for the visual adapter.} The results of other variants that activate text adapter are presented in Section~\ref{sec:text-adapter}. We use the same training hyperparameters as CoOp, including a batch size of $32$ and a learning rate of  $1\times10^{-5}$ for all datasets except for the residual ratio
$\alpha$. We perform hyperparameter searching over different value selections of $\alpha$ for each dataset and report the best performance among all searching spaces. We use ResNet-50~\cite{he2016deep} as the visual backbone (visual encoder) and 12-layer Transformer as classifier weight generator (textual encoder). The hidden embedding dimensionality of both visual and text bottleneck layers is set to $256$, which is a quarter of the original embedding dimensionality. In contrast to the learnable continuous prompts in CoOp, simple hand-crafted hard prompts are utilized as the text inputs of CLIP-Adapter, which is the same as CLIP. For generic-category image datasets, such as ImageNet, we adopt ``a photo of a \{\textsc{class}\}'' as the hard prompt template. For fine-grained classification datasets, we specify its corresponding domain keyword in the template for a better performance, for instance, ``a centered \emph{satellite} photo of \{\textsc{class}\}" for EuroSAT, and similarly for other fine-grained datasets.

\begin{figure*}[t!]
    \centering
    \begin{minipage}[t]{0.325\linewidth}
    \centering
    \includegraphics[width=1.57in]{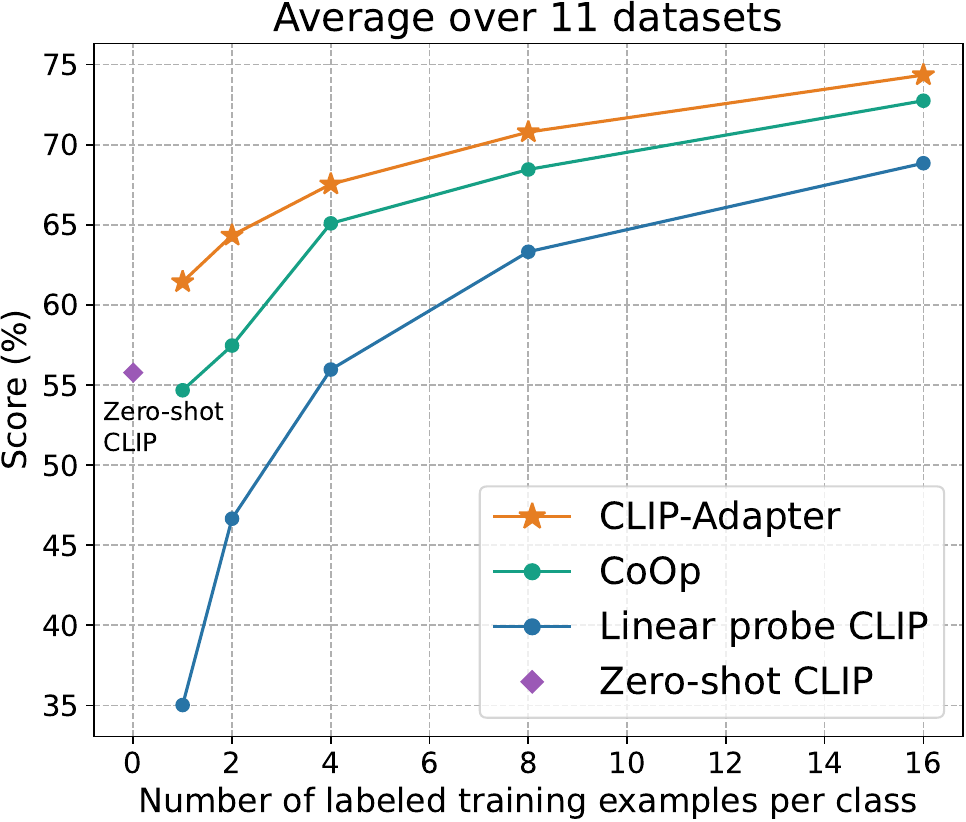}
    \end{minipage}
    \begin{minipage}[t]{0.325\linewidth}
    \centering
    \includegraphics[width=1.57in]{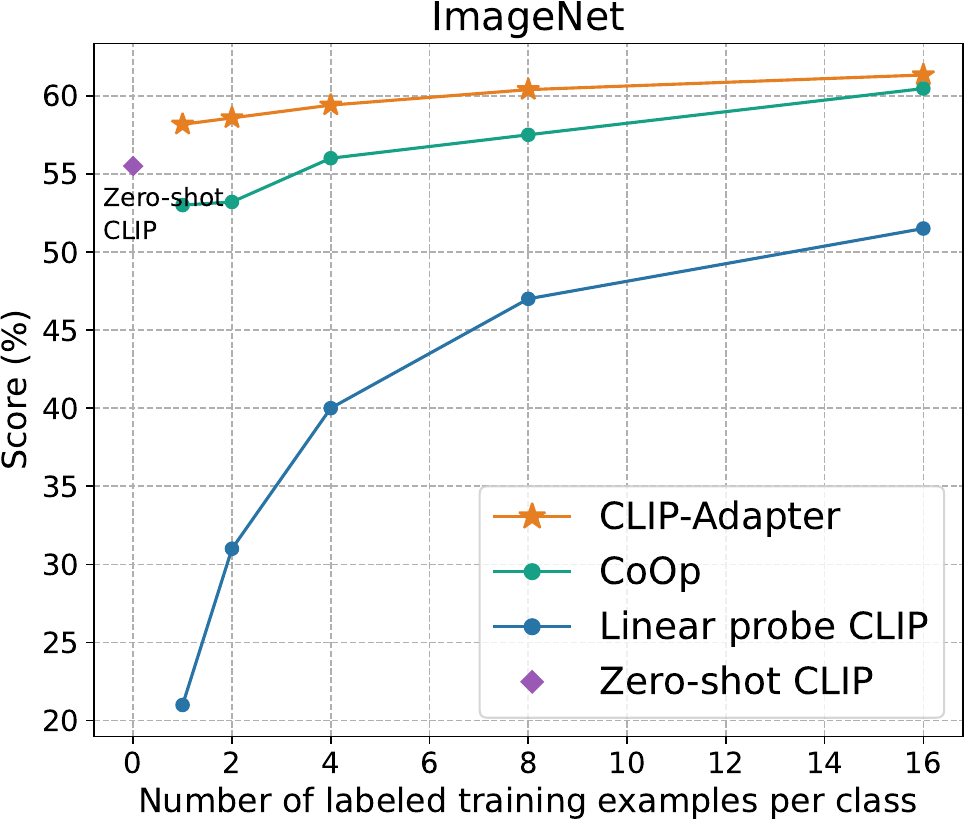}
    \end{minipage}
    \begin{minipage}[t]{0.325\linewidth}
    \centering
    \includegraphics[width=1.57in]{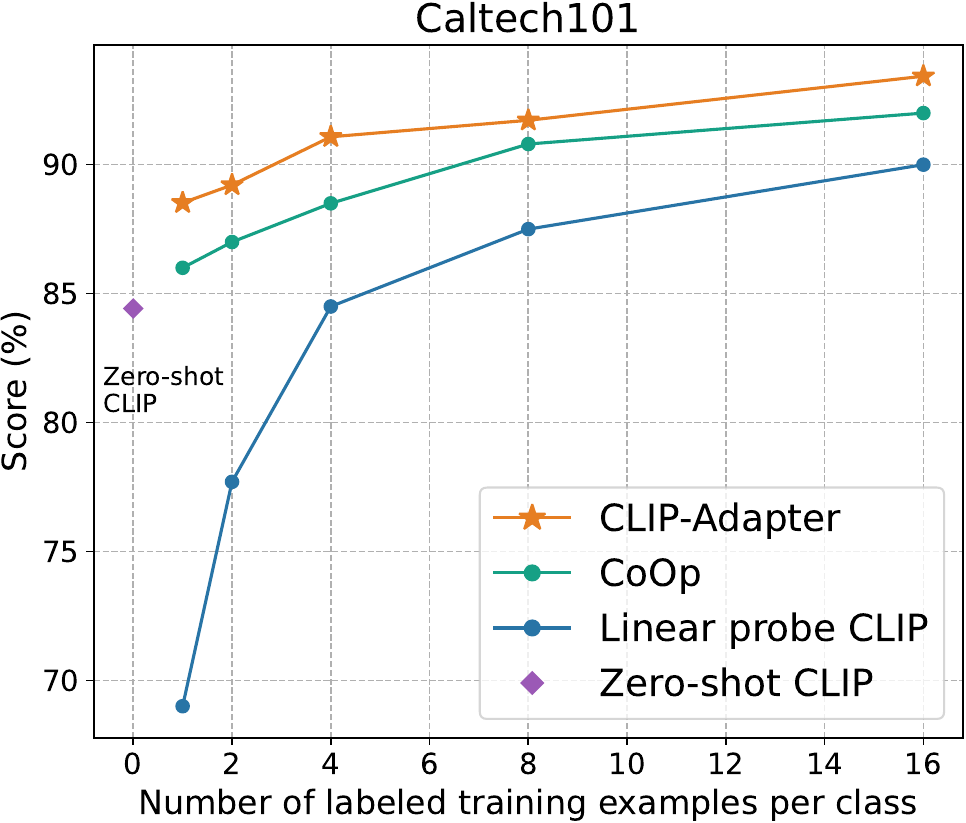}
    \end{minipage}
    
    \hspace{0.2in}
    
    \begin{minipage}[t]{0.325\linewidth}
    \centering
    \includegraphics[width=1.57in]{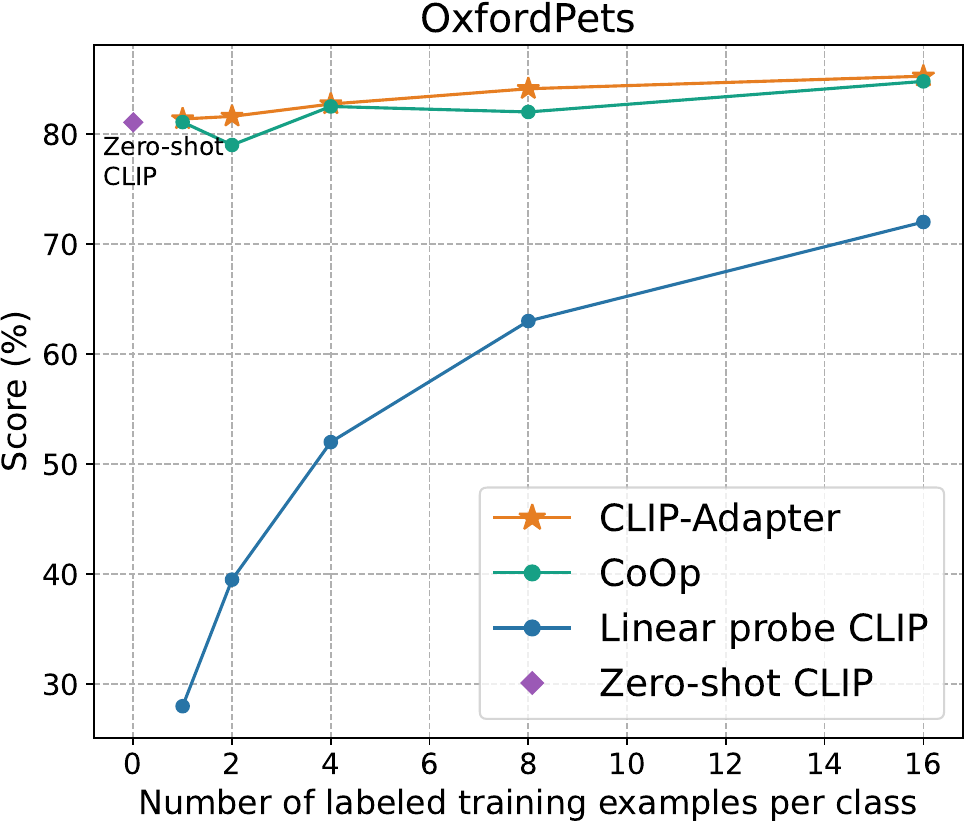}
    \end{minipage}
    \begin{minipage}[t]{0.325\linewidth}
    \centering
    \includegraphics[width=1.57in]{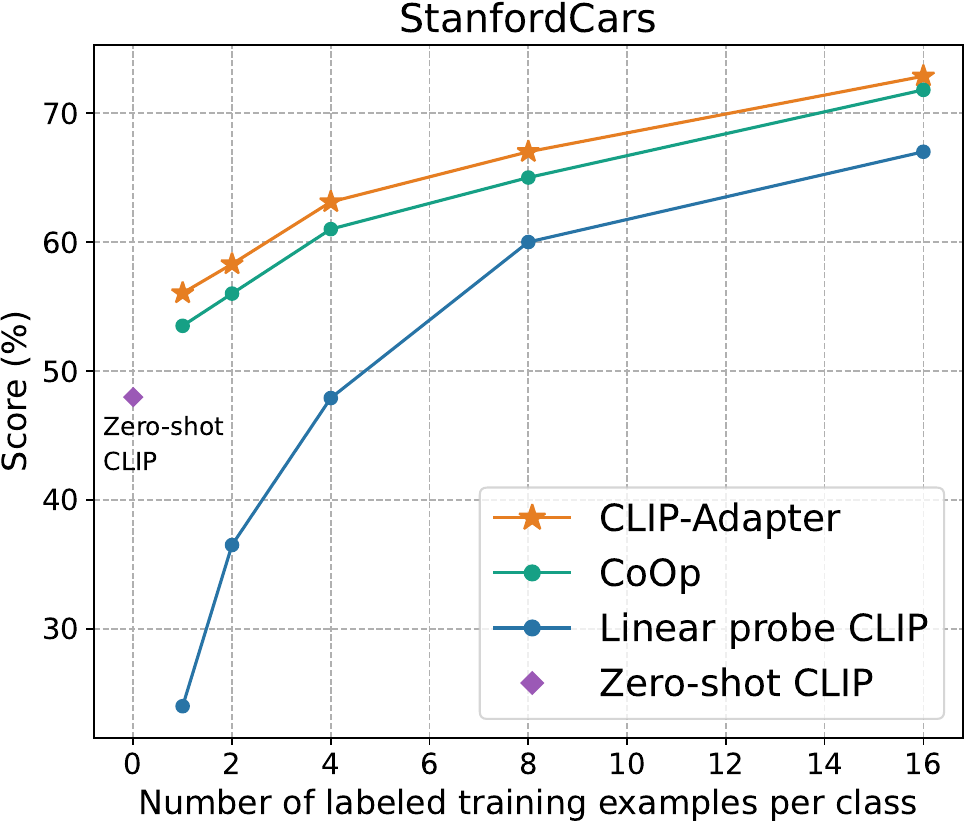}
    \end{minipage}
    \begin{minipage}[t]{0.325\linewidth}
    \centering
    \includegraphics[width=1.57in]{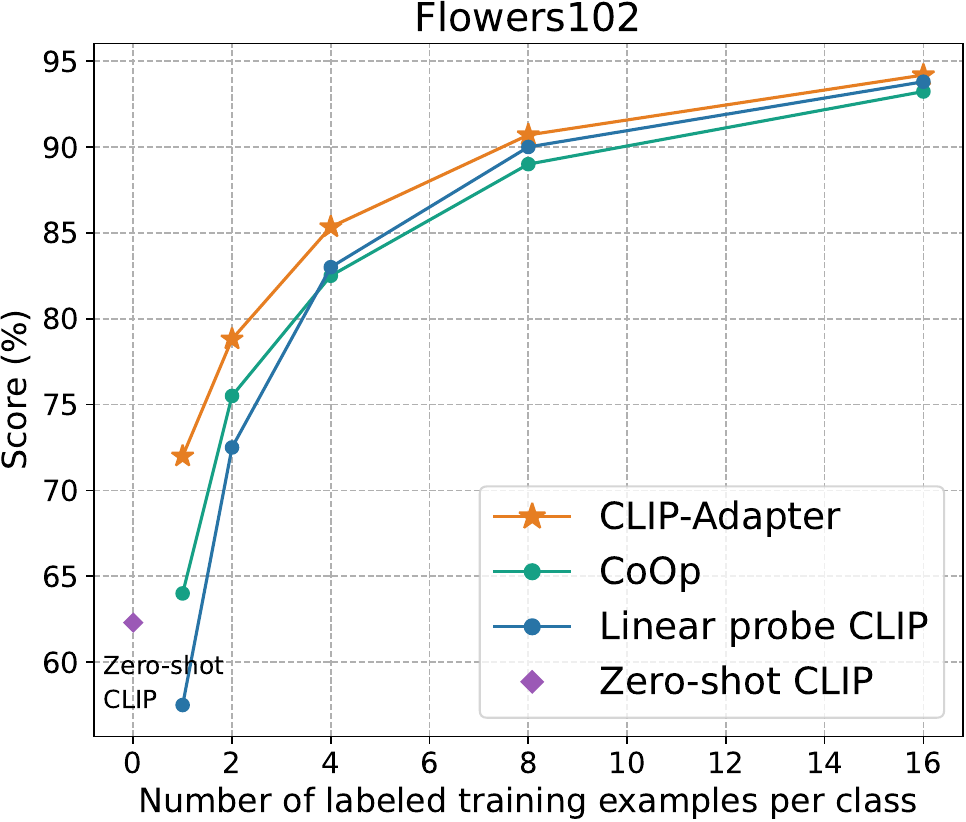}
    \end{minipage}
    
    \hspace{0.2in}
    
    \begin{minipage}[t]{0.325\linewidth}
    \centering
    \includegraphics[width=1.57in]{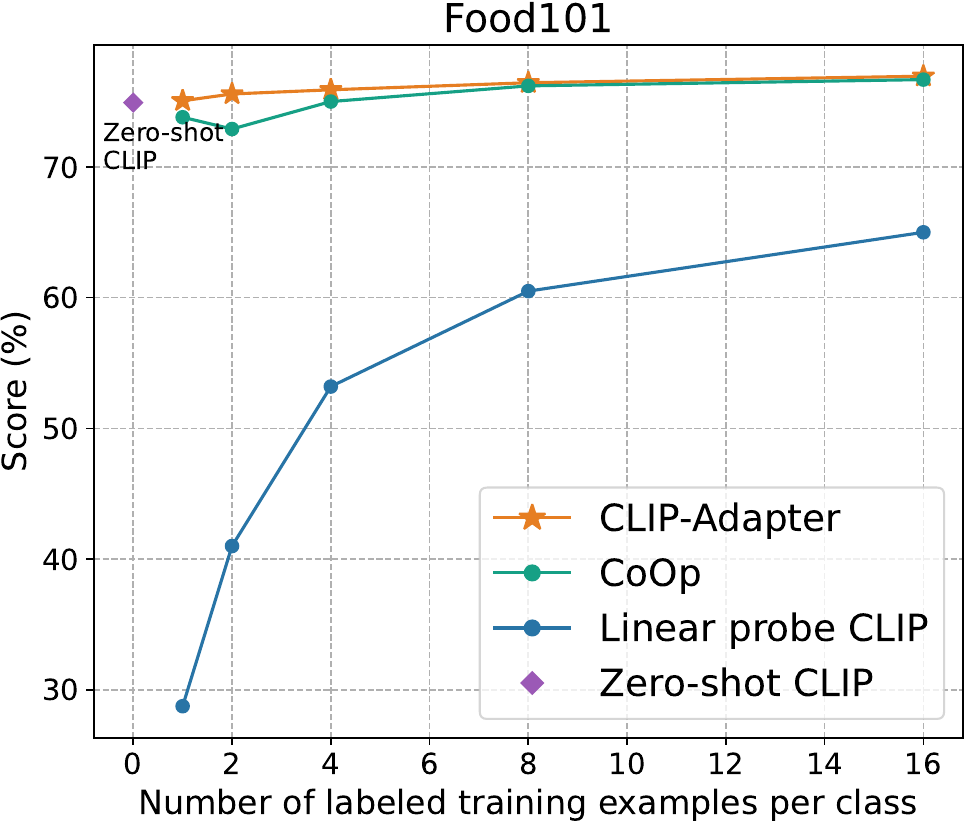}
    \end{minipage}
    \begin{minipage}[t]{0.325\linewidth}
    \centering
    \includegraphics[width=1.57in]{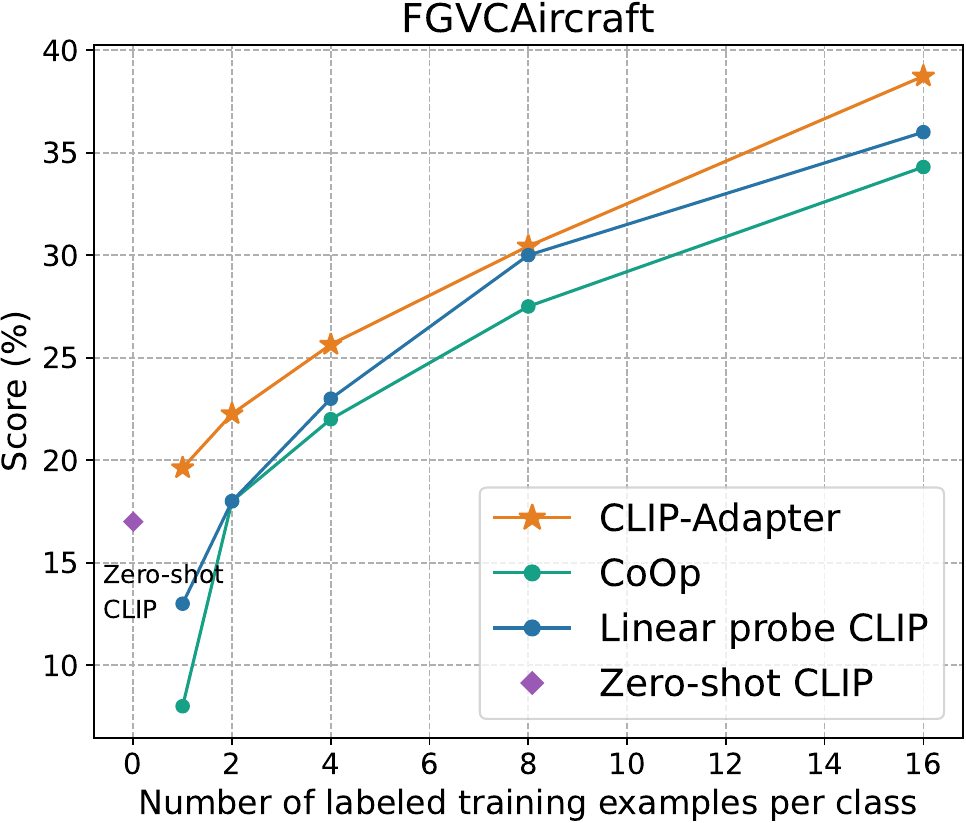}
    \end{minipage}
    \begin{minipage}[t]{0.325\linewidth}
    \centering
    \includegraphics[width=1.57in]{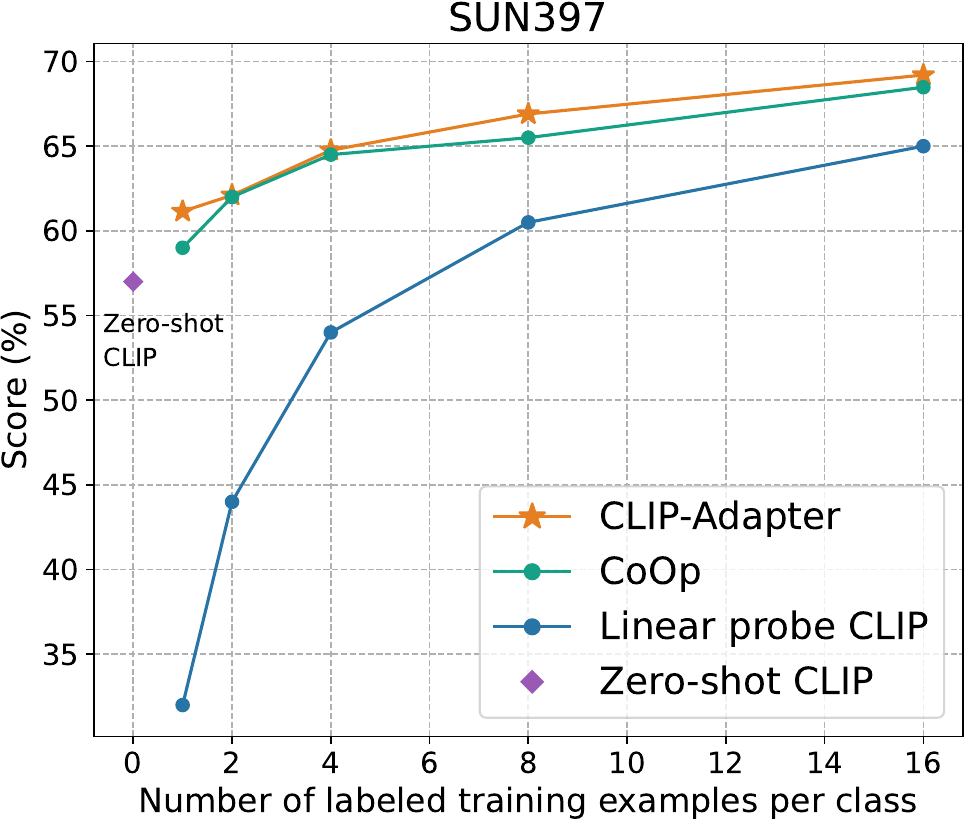}
    \end{minipage}
    
    \hspace{0.2in}
    
    \begin{minipage}[t]{0.325\linewidth}
    \centering
    \includegraphics[width=1.57in]{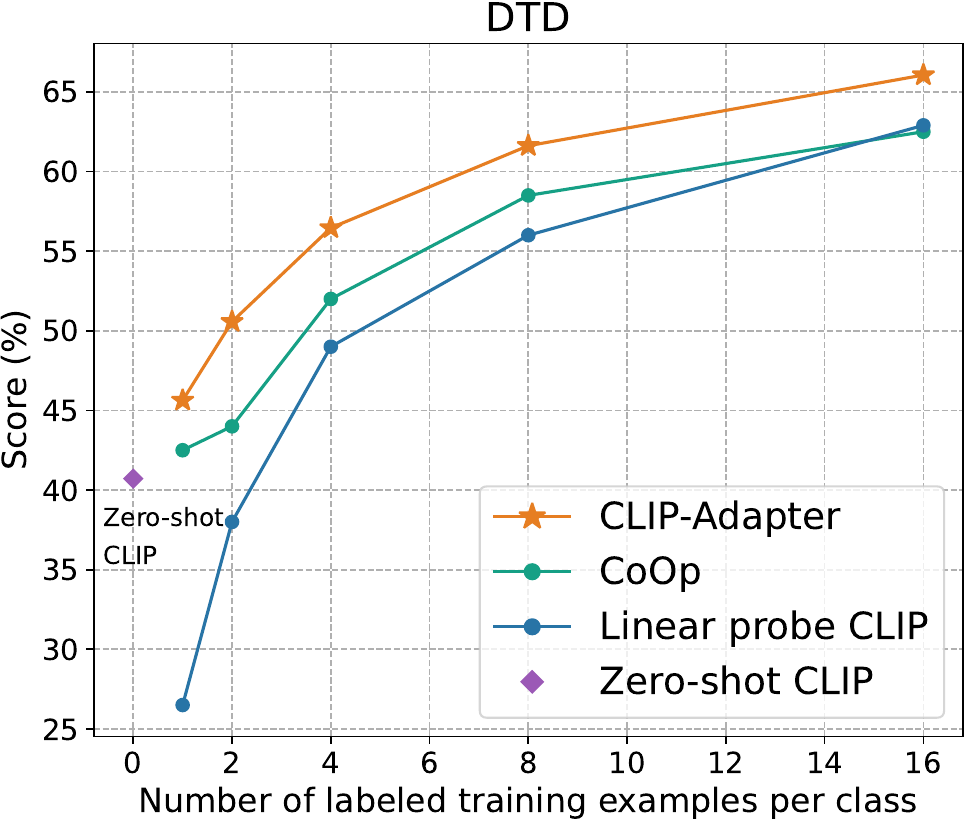}
    \end{minipage}
    \begin{minipage}[t]{0.325\linewidth}
    \centering
    \includegraphics[width=1.57in]{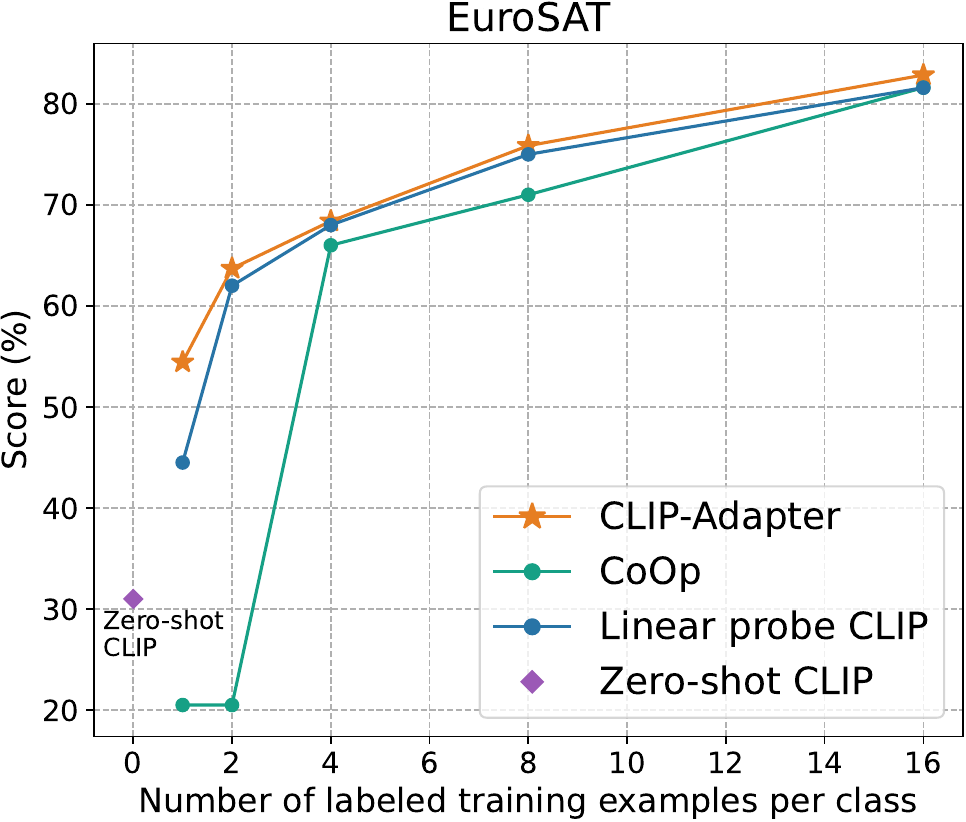}
    \end{minipage}
    \begin{minipage}[t]{0.325\linewidth}
    \centering
    \includegraphics[width=1.57in]{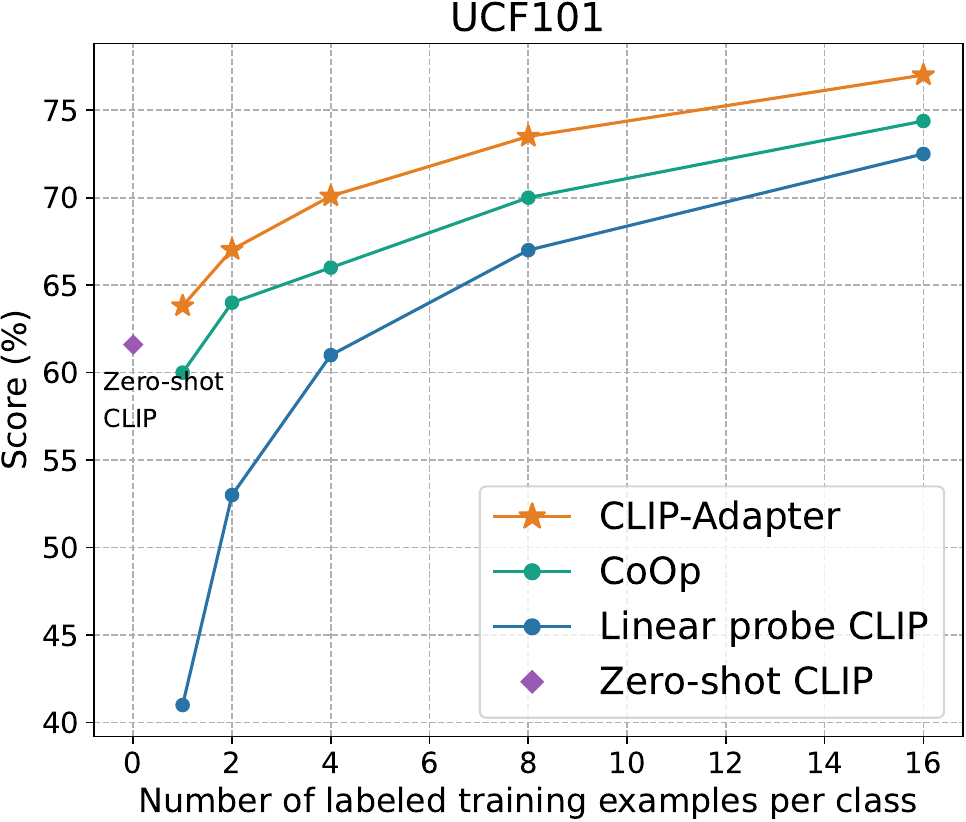}
    \end{minipage}
    \centering
    \caption{Main results of few-shot learning on 11 datasets. CLIP-Adapter consistently shows better performance over previous baselines across different training shots.}
    \label{fig:main_results}
\end{figure*}

\vspace{3pt}
\noindent\textbf{Notes on Image Pre-processing.~}
There are two image pre-processing methods adopted by existing methods. The first one is adopted by CLIP and the second one is reported in CoOp. We denote them as CLIP-style and CoOp-style preprocessings, respectively. They are both composed of random cropping, resizing, and random horizontal flip transformations. Their differences lie in the resizing. The CLIP-style pre-processing resizes the cropped image’s short side to $224$ while keeping its original aspect ratio. In contrast, the CoOp-style resizes an image's both sides to $224$. By default, we follow CoOp-style preprocessing. In Section~\ref{sec:appendix-clip-preprocessing} of the Appendix, we present the result comparison under the CLIP-style preprocessing which preserves the original aspect ratios of the cropped images.

\subsection{Comparison on Few-Shot Learning}

\subsubsection{Baseline Models}
We compare our CLIP-Adapter with three baseline models -- the \textbf{Zero-shot CLIP}~\cite{radford2021learning}, \textbf{Linear probe CLIP}~\cite{radford2021learning}, and \textbf{CoOp}~\cite{zhou2021coop}. In our implementation, CLIP-Adapter shares the same hand-crafted hard prompts with Zero-shot CLIP~\cite{radford2021learning} for fair comparisons. CoOp~\cite{zhou2021coop} substitutes discrete tokens with learnable continuous vectors. Thus there are multiple candidate positions to place the class token in the prompt template, namely at the front, in the middle, or at the end. Here, we choose CoOp's best-performance variant -- placing the class token at the end of the $16$-token soft prompt and shares such a context among different classes.~Linear probe CLIP~\cite{radford2021learning} trains an additional linear classifier on top of its visual encoder and follows a few-shot training manner. It is different from our bottleneck adapter that finetunes both the image feature and classifier weight in a dynamic and residual fashion.

\begin{table*}[ht]
\centering
\caption{Comparison of training time, parameters and classification accuracy of 16-shot learning on ImageNet~\cite{deng2009imagenet} dataset. CLIP-Adapter shows the best cost-accuracy trade-off. We test Linear Probe CLIP with Scikit-learn on CPU following CoOp, and implement other methods with batch size 32 and adopt ResNet-50 as the image backbone on a single NVIDIA A100 GPU.}
\vspace*{5pt}
\begin{adjustbox}{width=0.8\linewidth}
	\begin{tabular}{lccccccc}
	\toprule
		Models & Train Time & Parameters &GPU Mem. & Infer Speed & Accuracy\\ 
		\midrule
		Zero-shot CLIP  & 0 & 0 &2227 MiB &10.2ms & 55.41\%\\
		Linear Probe CLIP & 13min & 1.02M &-&- & 53.44\%\\
		CoOp  & 14h\ 40min & 0.02M &7193 MiB &299.6ms & 60.46\%\\
		CLIP-Adapter & 50min & 0.52M &2227 MiB &10.6ms & 61.33\% \\
	\bottomrule
	\end{tabular}
\end{adjustbox}
\label{time}
\end{table*}


\subsubsection{Performance Comparison \& Analysis}
The main results are presented in Figure~\ref{fig:main_results}. From the average accuracy over the 11 datasets shown at the top-left corner, CLIP-Adapter clearly outperforms the other three baseline models on all different shot setups, demonstrating its superior few-shot learning capacity. It is especially worth noticing that, under extreme conditions such as $1$-shot or $2$-shot training setup, CLIP-Adapter achieves larger performance improvements against the baselines, which indicates a better generalization ability in data-deficient training circumstances.

\begin{figure}[t!]
	\centering
	\includegraphics[height=5.3cm]{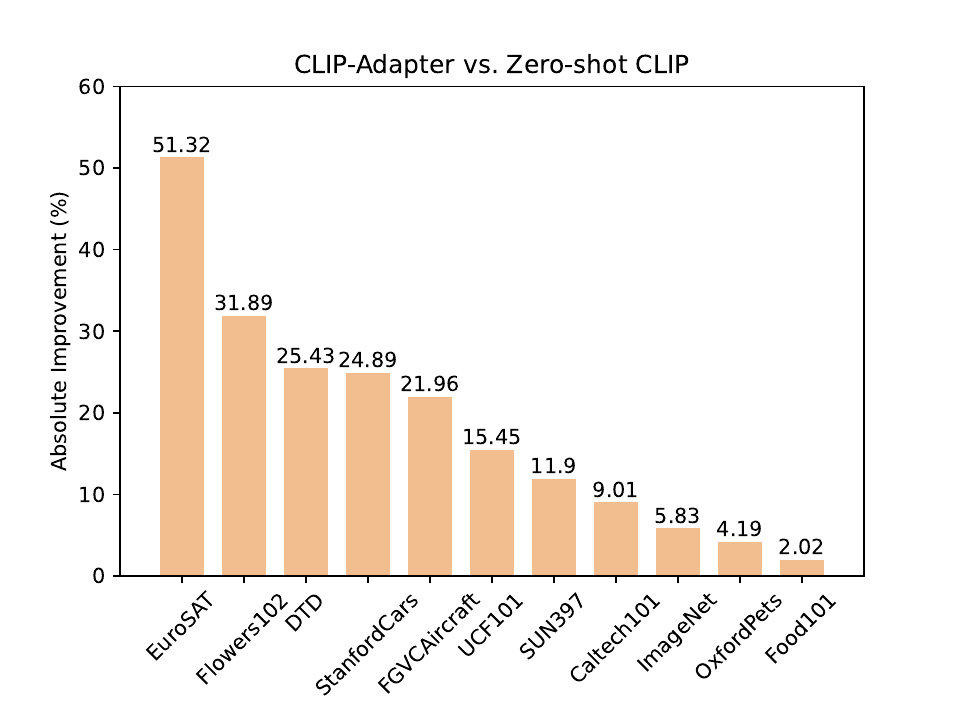}
	\caption{Absolute performance gain of CLIP-Adapter against hand-crafted prompts on different datasets.}
	\label{fig:vszero}
\end{figure}

Compared with \textbf{Zero-shot CLIP}~\cite{radford2021learning}, our CLIP-Adapter achieves significant performance gains over all 11 datasets.~The ranked absolute performance improvements for all 11 datasets under the 16-shot training setup are shown in Figure~\ref{fig:vszero}. For the first five fine-grained datasets, from EuroSAT to FGVCAircraft, CLIP-Adapter achieves huge performance boosts ranging from $20\%$ to $50\%$.~The improvements become smaller on more challenging and generic datasets, such as Caltech101 and ImageNet. As for OxfordPets and Food101,~CLIP-Adapter shows relatively limited improvements, since the original results of Zero-shot CLIP are already quite decent.

Compared with \textbf{Linear probe CLIP}~\cite{radford2021learning}, which follows a similar style to finetune the pretrained vision-language models, CLIP-Adapter also shows comprehensive performance advantages. Under 1-shot and 2-shot training setups, Linear probe CLIP barely reaches the performance of Zero-shot CLIP, but CLIP-Adapter can always surpass Zero-shot CLIP and exceed Linear probe CLIP by a large margin. For instance, the absolute margin of 1-shot and 2-shot training setups are $53.6\%$ and $42.16\%$ for OxfordPets, and $37.17\%$ and $27.58\%$ for ImageNet, respectively.

Compared with \textbf{CoOp}~\cite{zhou2021coop},~although it has already gained huge improvements over Zero-shot CLIP, CLIP-Adapter still outperforms CoOp on all datasets and different shot settings.~Note that CLIP-Adapter handles few-shot learning from a totally different perspective (i.e., fine-tuning) instead of CoOp's prompt tuning.~This suggests finetuning lightweight adapters with residual connections for prompt-fixed pretrained vision-language models can achieve better performance than prompt engineering~\cite{liu2023pre}.

\subsubsection{Efficiency Comparison \& Analysis}
\label{sec:efficiency}
In Table~\ref{time}, we provide the comparison of parameters, training budget, and inference speed for different methods. Compared to the baselines, CLIP-Adapter achieves the best accuracy while maintaining the best trade-off between parameters and efficiency. Specifically, CLIP-Adapter uses only half amount of parameters as Linear probe CLIP does. Although CLIP-Adapter costs 37 more minutes than Linear probe CLIP, CLIP-Adapter achieves a large 7.89\% accuracy boost.~Moreover, CLIP-Adpater takes $16\times$ less training time and $29\times$ faster inference speed than CoOp. Although CoOp only uses a small number of parameters, its learnable prompts are equipped in front of the text encoder and require both the forward and backward propagation, as shown by green dotted lines in Figure~\ref{fig:back}. Calculating the weights and gradients for the large-scale text encoder cost much more GPU memory and training time. In contrast, our CLIP-Adapter utilizes the hand-crafted prompts and only back-propagates the gradients through the lightweight adapters, achieving superior computational efficiency.

\begin{table*}[ht!]
\centering
\caption{Comparison of adapters inserted into different layers of CLIP. We adopt ViT-B/16 as the image backbone for the visual adapter,} and report the 16-shot ImageNet performance.
\vspace*{5pt}
\begin{adjustbox}{width=0.7\linewidth}
	\begin{tabular}{lcccccccc}
	\toprule
		Insert Layers & All    & 12  &  10  &  8 & 6 & 4 & 2 &0 \\
		\midrule
		Accuracy (\%)  & 67.67		&70.88		&\textbf{71.85}	&70.55		&70.03		&70.14	   	&69.43		&69.05 \\
		GPU Memory (GiB)  & 5.98	 &	\textbf{2.22}		 &2.65	 &	2.89	 &	3.23		 &4.46		 &4.88		 &5.14 \\
            Parameters (M)  & 5.20	&	\textbf{0.52}		&0.78		&0.78	&	0.78		&0.78		&0.78		&0.78 \\
	\bottomrule
	\end{tabular}
\end{adjustbox}
\label{tab:insert}
\end{table*}

\subsubsection{Observation on Optimal Residual Ratio}
\label{sec:distr-alpha}
Interestingly, we observe the best residual ratio $\alpha$, to some extent, reflects the characteristics of different datasets under the ``pretrain-finetuning'' paradigm.~A larger semantic gap between pretrained and finetuning datasets requires CLIP-Adapter to learn a higher portion of knowledge from the newly adapted feature compared to the original CLIP's output, thus resulting in a larger optimal residual ratio, and vice versa. For fine-grained datasets on specialized domains, like EuroSAT of satellite images and DTD of detailed textures, the optimal residual ratio $\alpha$ is usually located within the range from $0.6$ to $0.8$. By contrast, the best $\alpha$ value of comprehensive and generic image datasets (e.g., Caltech-101 and ImageNet) is often around $0.2$.

\begin{figure}[t!]
\centering
	\includegraphics[height=5.0cm]{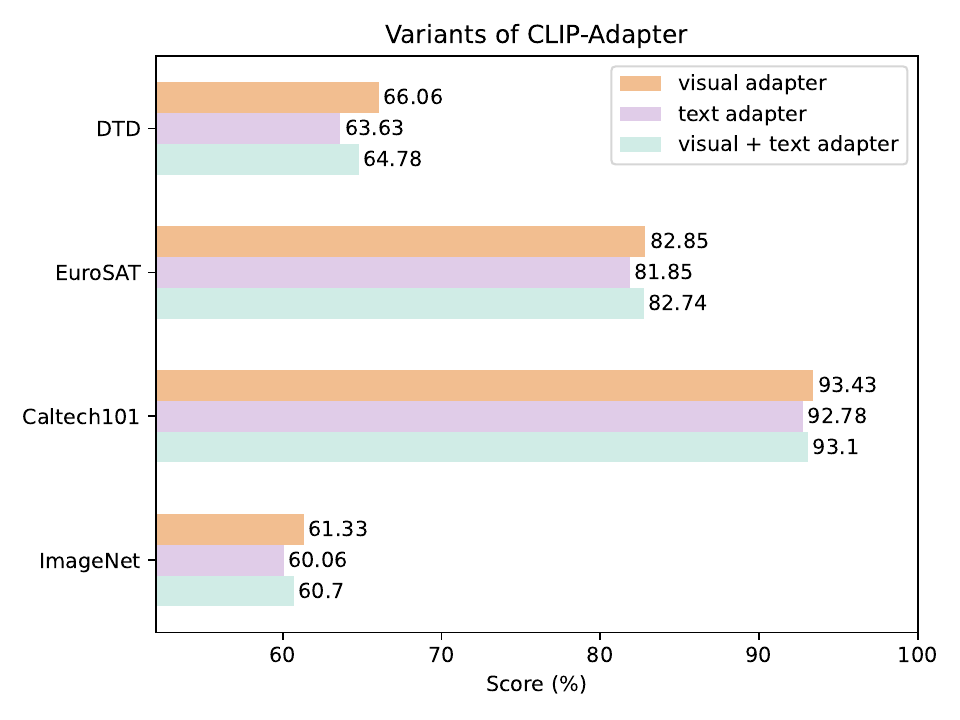}
	\caption{Comparison among different variants of CLIP-Adapter.}
	\label{fig:text_ad}
\end{figure}

\subsubsection{Variants with Text Adapter}
\label{sec:text-adapter}
Here, we investigate the other two variants of CLIP-Adapter mentioned in Section~\ref{sec:variants} -- finetuning the text adapter while keeping the visual adapter frozen and finetuning both the text and visual adapters. 
Rather than manually selecting the residual ratios for each dataset, we utilize learnable parameters $\alpha$ and $\beta$ since it is time-efficient and can also achieve satisfactory performance. We compare their performances on four datasets that can be divided into two categories -- fine-grained datasets (EuroSAT \& DTD) and generic datasets (Caltech101 \& ImageNet). As shown in Figure~\ref{fig:text_ad}, we can conclude that the text adapter and visual adapter perform comparably and both improve the classification accuracy greatly over Zero-shot CLIP. In addition, adopting visual adapter only is better than text adapter only. This indicates that it is more important to conduct image feature adaption than text feature adaption for few-shot image classification, since the semantic gap between visual features in pretrained and finetuning datasets is larger than that of text features. Surprisingly, combining both adapters together does not observe a better performance than visual adapter only.~This demonstrates that the text and visual adapters might capture redundant information or even conflict with each other.

\subsubsection{Where to Insert CLIP-Adapter?}
\label{sec:insert-adapter}
By default, we insert our residual-style adapters at the end of CLIP's encoder. We also investigate other positions in the visual backbone to equip our adapter. In Table~\ref{tab:insert}, we adopt ViT-B/16 as the visual backbone and respectively add the visual adapter after its 2nd, 4th, 6th, 8th, 10th, and 12th layers, where the 12th-layer variant denotes our final solution. As shown, our approach achieves superior performance with minimal computational cost when inserted at the end. Inserting at earlier layers requires more computation resources for back-propagating the gradients and, to some degree, harms the pretrained knowledge in CLIP. Inserting adapters in all layers yields a total of 5.20M parameters, which is much heavier-weight than inserting only at the 12th layer (0.52M). The latter can well alleviate the over-fitting issue on few-shot training data, and largely preserve the pre-trained knowledge of CLIP by inserting the adapter at the end.

\subsubsection{Comparison with other Adapter Methods}
\label{sec:other-adapter}
In Table~\ref{other-adapter}, we equip CLIP with other existing adapter-based methods~\cite{houlsby2019parameter,he2022towards} and compare with our CLIP-Adapter. As shown, our approach outperforms them in terms of both performance and efficiency. This is because we largely preserve the knowledge obtained from CLIP pretraining by inserting adapters at the end with residual connections, while others adopt non-residual forms and insert them densely in the middle of the backbone, which adversely influences CLIP's pretrained knowledge and results in overfitting.

\begin{table}[t!]
\centering
\caption{Comparison of CLIP-Adapter and existing adapter-based methods on ImageNet with the 16-shot setup. We adopt ViT-B/16 as the backbone with visual adapters.}
\vspace*{6pt}
\begin{adjustbox}{width=\linewidth}
	\begin{tabular}{lccc}
	\toprule
		Method &  CLIP-Adapter  &  Houlsby~\cite{houlsby2019parameter}  &  He~\cite{he2022towards}  \\
		\midrule
		Accuracy (\%)  & \textbf{70.88}		&70.16				&70.84 \\
  GPU Memory (GiB) &\textbf{2.22}			&6.74				&6.06\\
  Parameters (M) &\textbf{0.52}			&12.84				&9.62\\
	\bottomrule
	\end{tabular}
\end{adjustbox}
\label{other-adapter}
\end{table}

\begin{table}[t!]
\centering
\caption{Comparison of CLIP-Adapter and ELEVATER benchmark approaches on ImageNet with the 16-shot setup. We adopt ViT-B/16 as the backbone with visual adapters.}
\vspace*{6pt}
\begin{adjustbox}{width=\linewidth}
	\begin{tabular}{lccccc}
	\toprule
 \multirow{2.5}{*}{Method} &\multicolumn{3}{c}{ELEVATER~\cite{li2022elevater}} &\multirow{2.5}{*}{CoOp} &\multirow{2.5}{*}{CLIP-Adapter}\\\cmidrule{2-4}
		&R-2P &L-2P &L-1P &\\
  \midrule
		Accuracy (\%)  & 68.94		&69.77				&70.38 &70.16				&\textbf{70.88}\\
	\bottomrule
	\end{tabular}
\end{adjustbox}
\label{ELEVATER}
\end{table}

\subsubsection{Comparison with ELEVATER Benchmark Baselines}
\label{sec:ELEVATER}
In Table~\ref{ELEVATER}, we compare CLIP-Adapter with the newly proposed ELEVATER benchmark~\cite{li2022elevater} for pretrained vision-language models. ELEVATER includes three baselines: Random-Init with Two-Projection, Language-Init with Two-Projection, and Language-Init with One-Projection. As shown in the table, compared to training additional projection layers initialized by randomness on the text encoder, our CLIP-Adapter achieves higher classification accuracy, since the residual design can largely preserve the pretrained knowledge in CLIP.

\begin{table*}[ht!]
\centering
\caption{Ablations on varying the hidden dimension of bottleneck layers.}
\vspace*{5pt}
\begin{adjustbox}{width=0.6\linewidth}
	\begin{tabular}{lcccccc}
	\toprule
		Dimension & $D$    &  $D/2$  &  $D/4$  &  $D/8$ & $D/16$ &  $D/32$ \\
		\midrule
		DTD (\%)  & 65.03 & 65.62 &  \textbf{66.06} & 64.93 & 63.75 & 63.50 \\
		ImageNet (\%)  & 59.78 &  60.03 & \textbf{61.33} & 60.06 & 60.02 & 59.45 \\
	\bottomrule
	\end{tabular}
\end{adjustbox}
\label{dimension}
\end{table*}

\begin{table*}[ht]
\centering
\caption{Ablations on varying the residual ratio $\alpha$.}
\vspace*{5pt}
\begin{adjustbox}{width=0.6\linewidth}
	\begin{tabular}{lcccccc}
	\toprule
		Ratio $\alpha$ & $0$    & $0.2$  & $0.4$ & $0.6$ & $0.8$ & $1.0$ \\
		\midrule
		DTD (\%)  & 40.72 & 54.59 & 64.84 & \textbf{66.06} & 65.96 & 63.79 \\
		ImageNet (\%)  & 60.46 & \textbf{61.33} & 61.17 & 60.77 & 59.79 & 59.05 \\
	\bottomrule
	\end{tabular}
\end{adjustbox}
\label{alpha}
\end{table*}

\begin{figure*}[t!]
\centering
\includegraphics[width=0.92\linewidth,trim={0cm 0cm 0cm 0cm}]{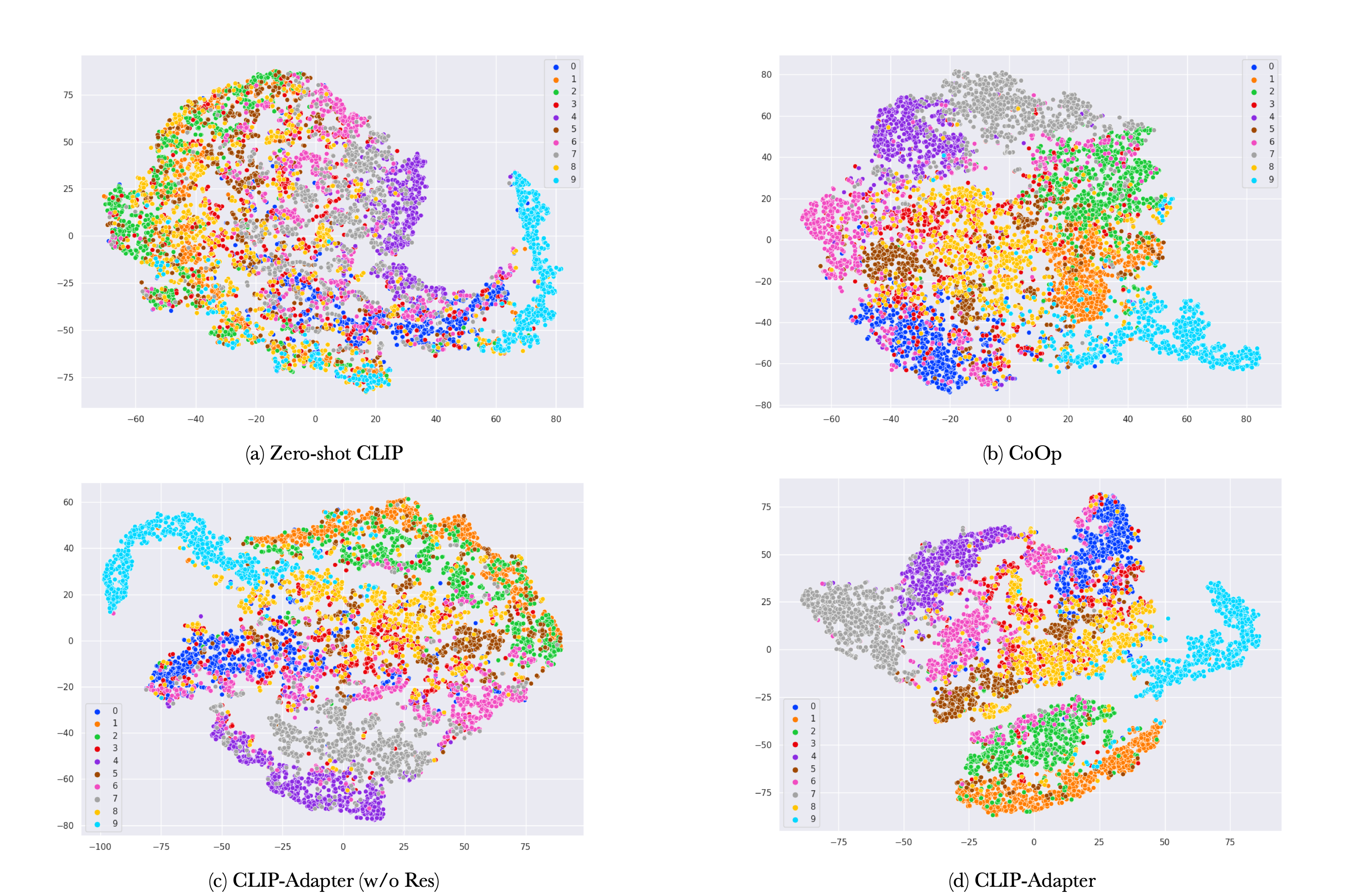}
\caption{Visualization of different learned feature manifolds via t-SNE.}
\label{fig:manifold}
\end{figure*}

\subsection{Visualization of Learned Manifold}
We use t-SNE~\cite{van2008visualizing} to visualize the manifold of CLIP, CoOp, CLIP-Adapter without residual connections, and CLIP-Adapter with residual connections after training them on the EuroSAT dataset.~The t-SNE visualization results are presented in Figure~\ref{fig:manifold}, where the numbers $0$ to $9$ stand for the categories of \textit{AnnualCrop}, \textit{Forest}, \textit{Herbaceous Vegetation Land}, \textit{Highway or Road}, \textit{Industrial Buildings}, \textit{Pasture Land}, \textit{Permanent Crop Land}, \textit{Residential Buildings}, \textit{River}, \textit{Sea or Lake}, respectively. It is clearly illustrated that in high-dimensional classification space, the CLIP-Adapter with residual connections in sub-figure (d) shows much more obvious separation of image features belong to different categories. As for the confusing categories such as \textit{Highway or Road}~(red points), \textit{Permanent Crop Land}~(pink points), and \textit{Pasture Land}~(brown points), compared with other methods, our CLIP-Adapter is more effective in detecting the similarities among the image manifolds from the same class. In summary, the visualization results prove that CLIP-Adapter is good at learning better feature manifolds under few-shot setups.

\subsection{Ablation Studies}
In this section, we perform several ablation studies for CLIP-Adapter.~We choose the best-performance variant which only activates the visual adapter, and select two datasets -- DTD \& ImageNet, serving as the representatives of fine-grained and generic datasets, to perform the ablation studies.

\begin{table}[ht!]
\centering
\caption{Few-shot performance on ImageNet using different prompt styles.}
\vspace*{6pt}
\begin{adjustbox}{width=\linewidth}
	\begin{tabular}{lccc}
	\toprule
		Prompt Style &  Hard  &  Hard Ensemble  &  Hard + Soft  \\
		\midrule
		CLIP-Adapter (\%)  & 61.33 & \textbf{61.68} & 59.69 \\
	\bottomrule
	\end{tabular}
\end{adjustbox}
\label{prompt}
\end{table}

\vspace{3pt}
\noindent\textbf{Dimension of Bottleneck Layer.~} We first conduct ablations by varying the hidden dimension of bottleneck layers.~The results are shown in Table~\ref{dimension}, where $D$ represents the dimension of the original image feature. By reducing the hidden dimension from $D$ to $D/32$, we observe that either too small or too large intermediate dimensionality will deteriorate the performance significantly and the best bottleneck dimension is $D/4$, which is able to preserve enough semantics without redundancy.

\begin{table*}[ht]
\centering
\caption{Ablations of different visual backbones.}
\vspace*{5pt}
\begin{adjustbox}{width=0.7\linewidth}
	\begin{tabular}{lccccc}
	\toprule
		Dataset & Method & RN50   & RN101  & ViT-B/32  & ViT-B/16 \\
		\midrule
		\multirow{2}*{DTD(\%)} & CoOp   & 62.55 & 65.37 & 65.43 & 67.67 \\
		& CLIP-Adapter  & \textbf{66.06} & \textbf{67.02} & \textbf{66.37} &\textbf{70.86} \\
		\midrule
		 \multirow{2}*{ImageNet(\%)} & CoOp   & 60.46 & 64.39 & 64.92 & 70.13 \\
		& CLIP-Adapter  & \textbf{61.33} & \textbf{64.77} & \textbf{64.99} & \textbf{70.88} \\
	\bottomrule
	\end{tabular}
\end{adjustbox}
\label{backbone}
\end{table*}

\begin{table*}[ht]
\centering
\caption{Evaluation on robustness to distribution shift.}
\vspace*{5pt}
\begin{adjustbox}{width=0.85\linewidth}
\begin{tabular}{lcccccc}
\toprule
\multirow{2.5}{*}{Datasets} & \textbf{Source} & \multicolumn{4}{c}{\textbf{Target}} \\
\cmidrule(lr){2-2} \cmidrule(lr){3-6} 
& ImageNet   & ImageNetV2 & ImageNet-Sketch & ImageNet-A &  ImageNet-R  \\
\midrule
Zero-Shot CLIP  & 55.41  & 48.08 & 31.67 & 18.63 & 53.45  \\
Linear Probe CLIP  & 53.44 & 43.40 & 17.63 & 11.66 & 32.63  \\
\midrule
CoOp & 60.46  & 52.17 & 31.14  &  19.62 &  53.31 \\
CLIP-Adapter & \textbf{61.33}  & \textbf{52.67} & \textbf{32.04} & \textbf{20.12} & \textbf{54.75} \\
\bottomrule
\end{tabular}
\end{adjustbox}
\label{tab:robust}
\end{table*}

\begin{table*}[t!]
\centering
\caption{Comparison of finetuning different components of CLIP-Adapter.}
\vspace*{5pt}
\begin{adjustbox}{width=0.7\linewidth}
	\begin{tabular}{ccccc}
	\toprule
		Visual Encoder &Textual Encoder
		 &Adapter &Accuracy & Train Time\\ \midrule
		- & - &\Checkmark &\textbf{61.33\%} &\textbf{50min}\\
		\Checkmark & - &\Checkmark &60.07\% &1h 20min\\
		- & \Checkmark &\Checkmark &57.88\% &3h+\\
	    \Checkmark & \Checkmark & \Checkmark & 52.78\% &3h+\\
	\bottomrule
	\end{tabular}
\end{adjustbox}
\label{tab:finetune}
\end{table*}

\vspace{3pt}
\noindent\textbf{Residual Ratio $\alpha$.~} Moreover, we perform ablation study of the residual ratio $\alpha$.~From Table~\ref{alpha}, we can see that the best residual ratio of fine-grained dataset DTD is $0.6$, and that of generic dataset ImageNet is $0.2$. This verifies our observation in Section~\ref{sec:distr-alpha} that adapting fine-grained dataset requires more new knowledge than old knowledge, and the case is opposite to the generic dataset. Note that when $\alpha$ equals to $0$, it is equivalent to Zero-shot CLIP since no new knowledge is learned. When $\alpha$ is set to $1.0$, the classification is fully rely on the adapted feature (CLIP-Adapter w/o Res).~However, this is not optimal because CLIP-Adapter tends to overfit in such condition.~Combining Table~\ref{alpha} and Figure~\ref{fig:manifold}, we can also conclude the advantages of residual connections in CLIP-Adapter:~1) avoids overfitting on few-shot examples and improves the generalization ability of CLIP-Adapter with the help of zero-shot knowledge;~2) preserves the freedom for learning better image feature or classifier weight through few-shot fine-tuning.

\vspace{3pt}
\noindent\textbf{Influence of Prompt Styles.~} In this section, we investigate the influence of different prompt styles on few-shot performance. For ImageNet dataset, the default hard prompt used as text inputs of CLIP-Adapter is simply ``a photo of a \{\textsc{class}\}''. Besides, we also try prompt ensembling~\cite{zhou2021coop} of 7 hard prompts. The 7 hard prompt templates are: ``itap of a \{\textsc{class}\}'', ``a bad photo of the \{\textsc{class}\}'', ``a origami \{\textsc{class}\}'', ``a photo of the large \{\textsc{class}\}'', ``a \{\textsc{class}\} in a video game'', ``art of the \{\textsc{class}\}'' and ``a photo of the small \{\textsc{class}\}''. Another candidate prompt style is the mixture of hard prompt and learnable soft prompt~\cite{zhou2021coop}. As shown in Table~\ref{prompt}, the prompt ensembling strategy slightly outperforms hard prompt and achieves the best performance among all three prompt styles. The experimental results prove that raw text descriptions contain helpful knowledge which is effective and robust under different situations. In contrast, soft prompts don't have clear meaning and are not a ideal source for zero-shot knowledge.

\vspace{3pt}
\noindent\textbf{Ablation of Visual Backbones.~} We also study the influence of visual backbones on few-shot learning performance (16 shots). There are four candidate visual backbones including ResNet-50, ResNet-101, ViT-B/32, and ViT-B/16. As reported in Table~\ref{backbone}, CLIP-Adapter consistently outperforms CoOp when we vary the visual backbones on both DTD and ImageNet datasets.

\vspace{3pt}
\noindent\textbf{Robustness under Distribution Shift.~} To further validate the robustness of CLIP-Adpater, we also perform experiments to observe performance variation by shifting the distribution. We train our CLIP-Adapter on ImageNet~\cite{deng2009imagenet} and respectively evaluate on four out-of-distribution datasets: ImageNetV2~\cite{recht2019imagenet}, ImageNet-Sketch~\cite{hendrycks2021natural}, ImageNet-A~\cite{wang2019learning}, and ImageNet-R~\cite{hendrycks2021many}. As shown in Table~\ref{tab:robust}, CLIP-Adapter consistently outperforms other baselines and demonstrates enough robustness against distribution shift.


\vspace{2pt}
\noindent\textbf{Finetuning Whole CLIP vs. CLIP-Adapter.~} To verify the claim that finetuning the whole CLIP would lead to overfitting. We perform ablation experiments on finetuning different components of CLIP-Adapter ($\checkmark$ denotes unfrozen for training). For finetuning CLIP's encoders, we adopt early stopping as suggested to obtain the highest accuracy. From the results presented in Table \ref{tab:finetune}, we observe that finetuning either CLIP's visual or textual encoder would hurt the performance and take more training time. This indicates the overfitting of the huge-parameter CLIP on the few-shot dataset and the effectiveness of the proposed adapter.

\begin{figure*}[t!]
    \centering
    \begin{minipage}[t]{0.325\linewidth}
    \centering
    \includegraphics[width=1.57in]{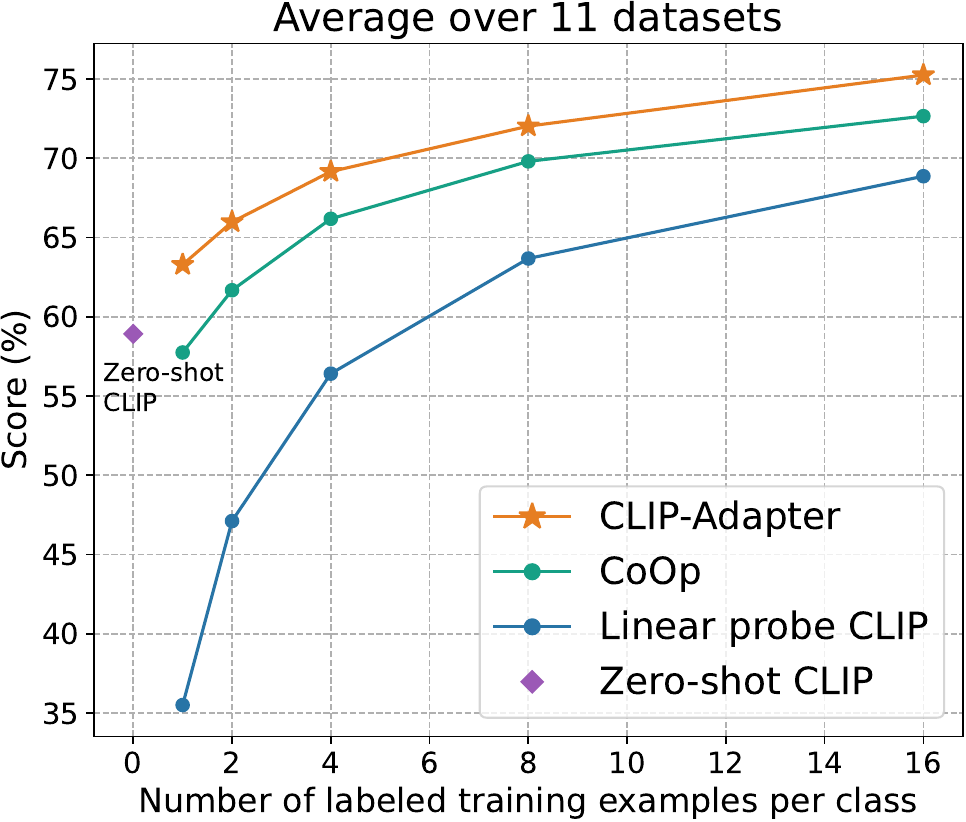}
    \end{minipage}
    \begin{minipage}[t]{0.325\linewidth}
    \centering
    \includegraphics[width=1.57in]{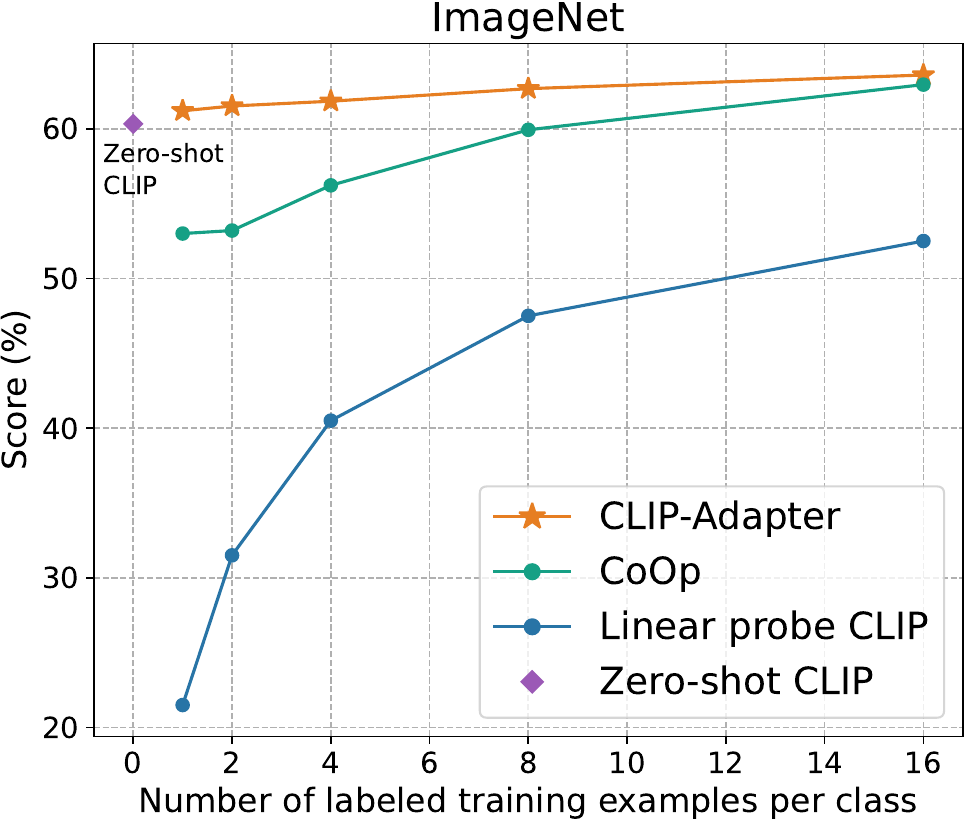}
    \end{minipage}
    \begin{minipage}[t]{0.325\linewidth}
    \centering
    \includegraphics[width=1.57in]{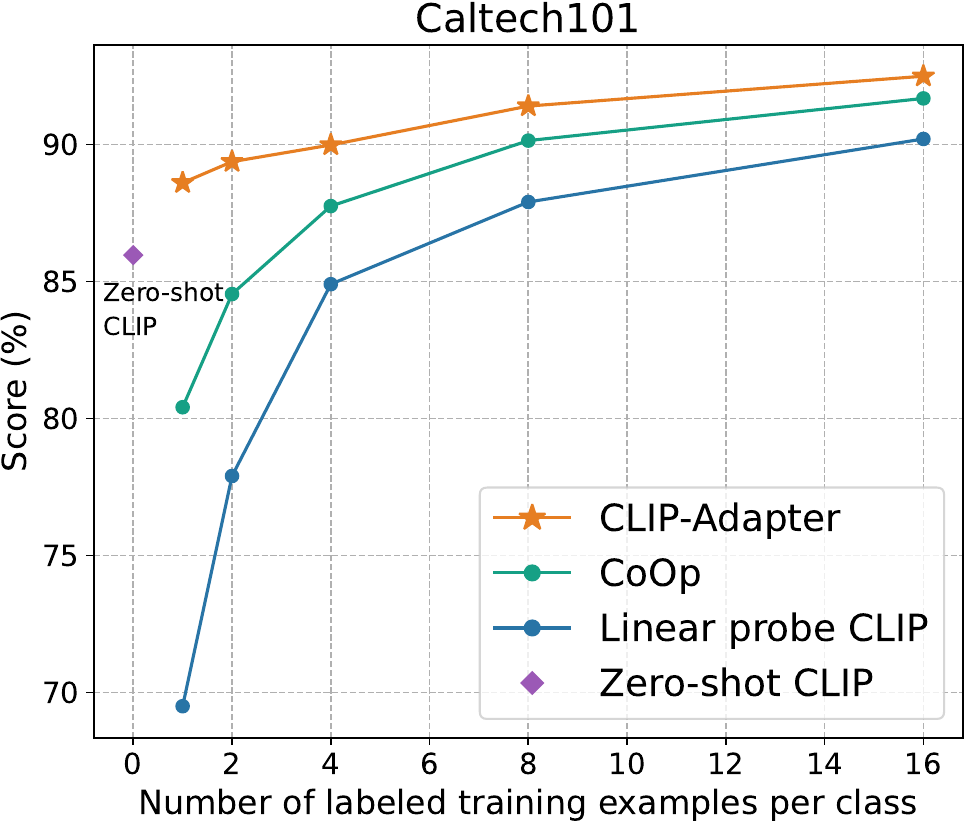}
    \end{minipage}
    
    \hspace{0.2in}
    
    \begin{minipage}[t]{0.325\linewidth}
    \centering
    \includegraphics[width=1.57in]{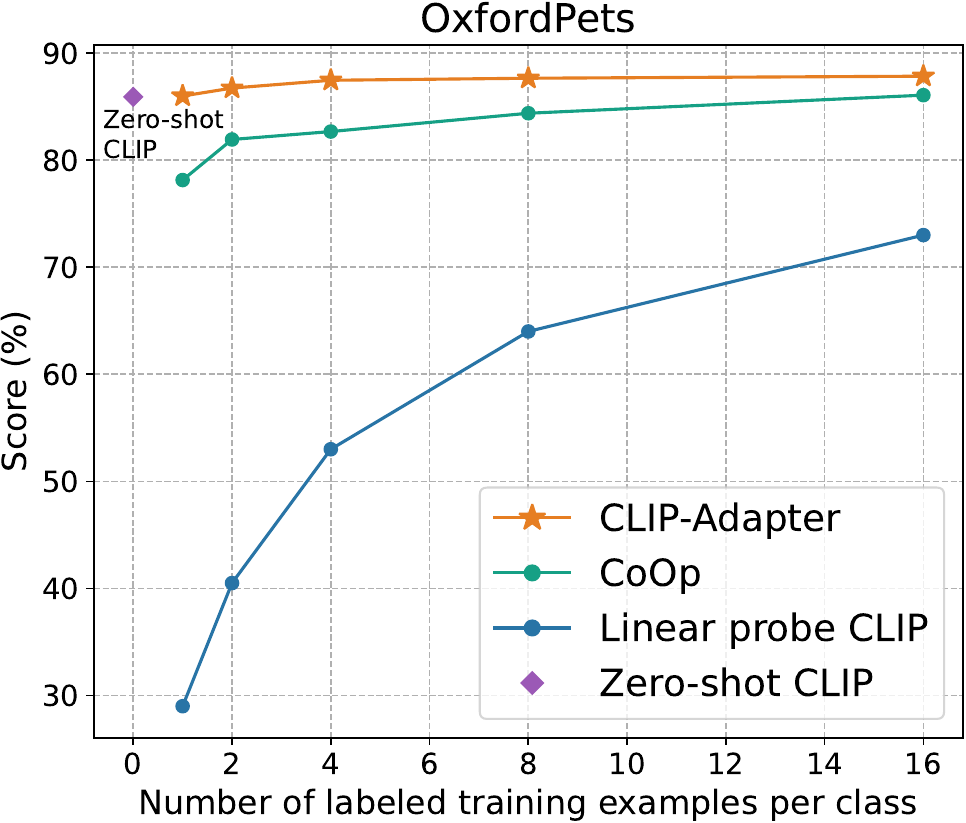}
    \end{minipage}
    \begin{minipage}[t]{0.325\linewidth}
    \centering
    \includegraphics[width=1.57in]{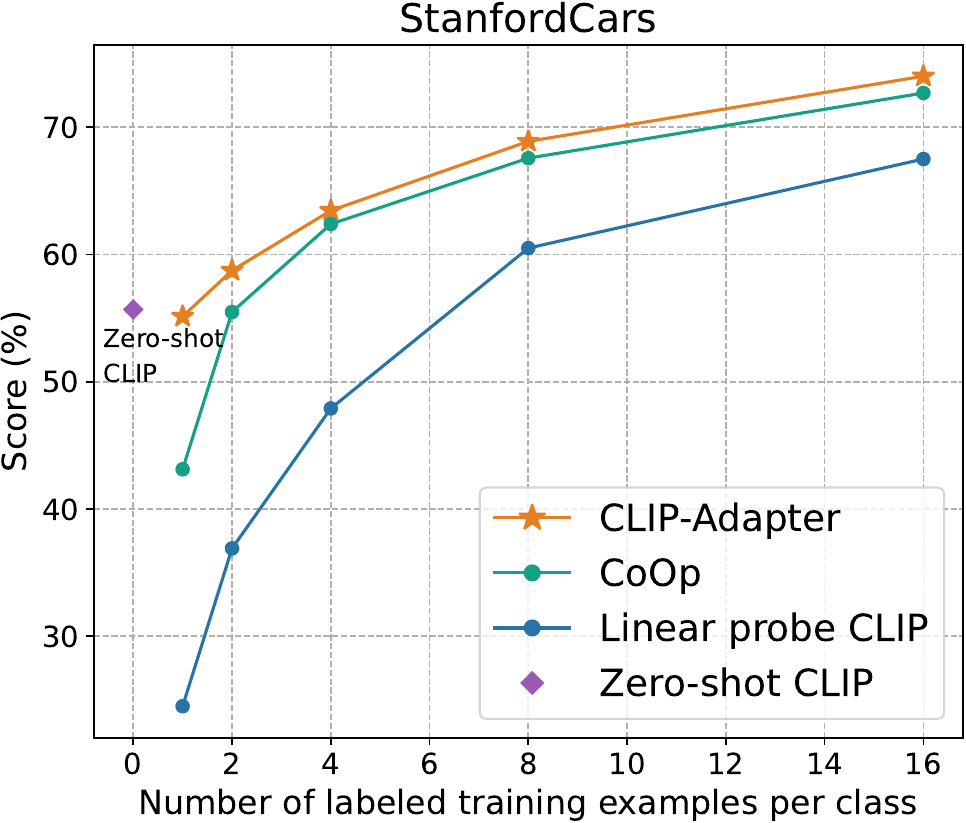}
    \end{minipage}
    \begin{minipage}[t]{0.325\linewidth}
    \centering
    \includegraphics[width=1.57in]{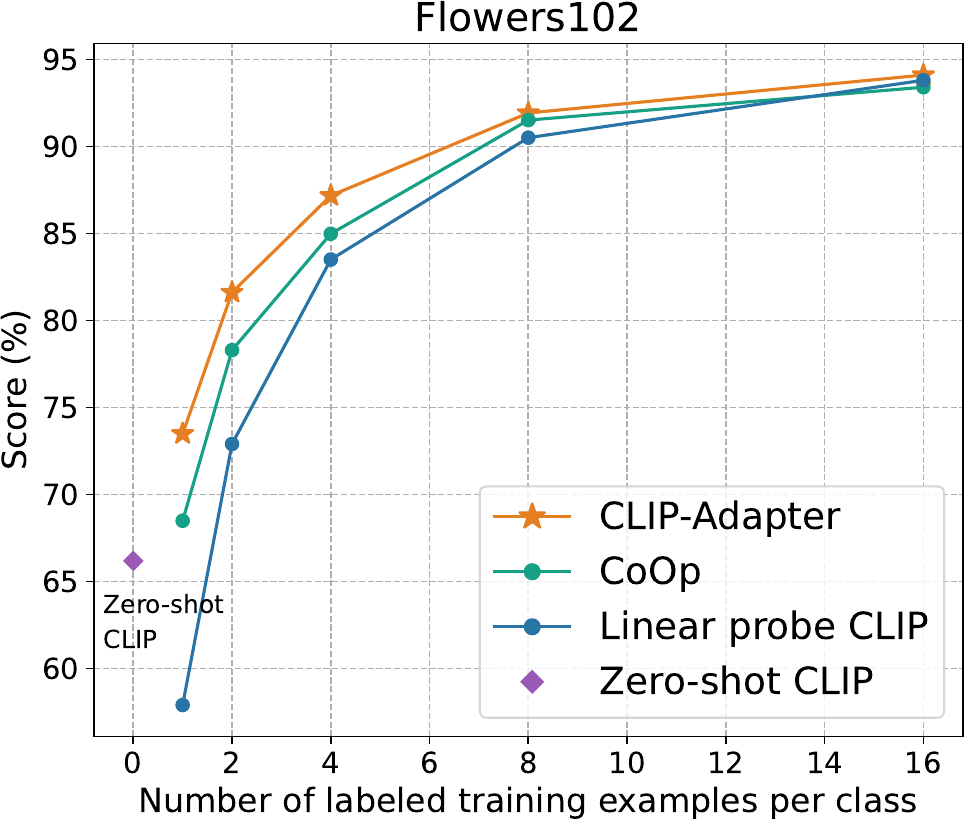}
    \end{minipage}
    
    \hspace{0.2in}
    
    \begin{minipage}[t]{0.325\linewidth}
    \centering
    \includegraphics[width=1.57in]{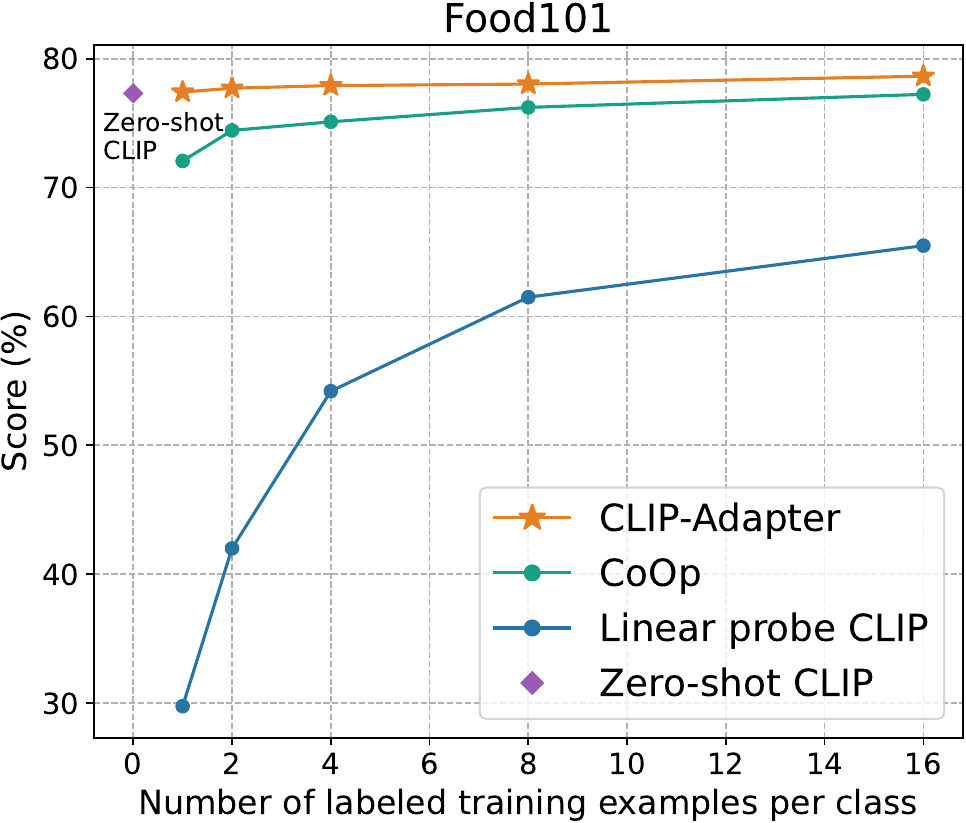}
    \end{minipage}
    \begin{minipage}[t]{0.325\linewidth}
    \centering
    \includegraphics[width=1.57in]{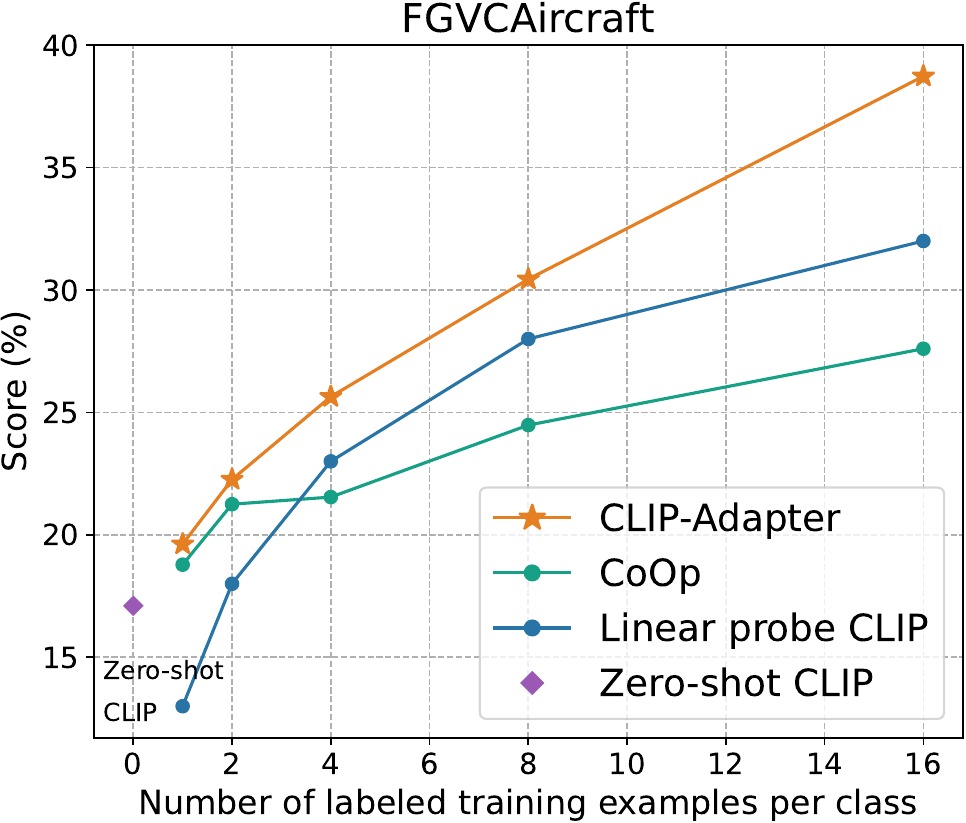}
    \end{minipage}
    \begin{minipage}[t]{0.325\linewidth}
    \centering
    \includegraphics[width=1.57in]{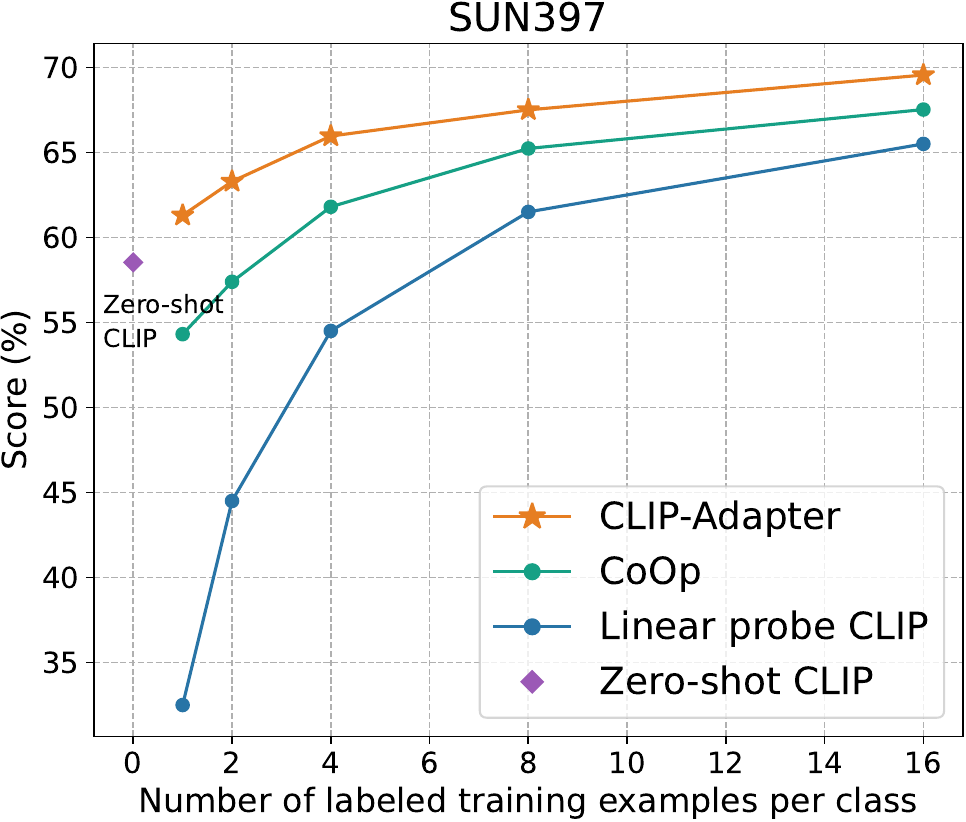}
    \end{minipage}
    
    \hspace{0.2in}
    
    \begin{minipage}[t]{0.325\linewidth}
    \centering
    \includegraphics[width=1.57in]{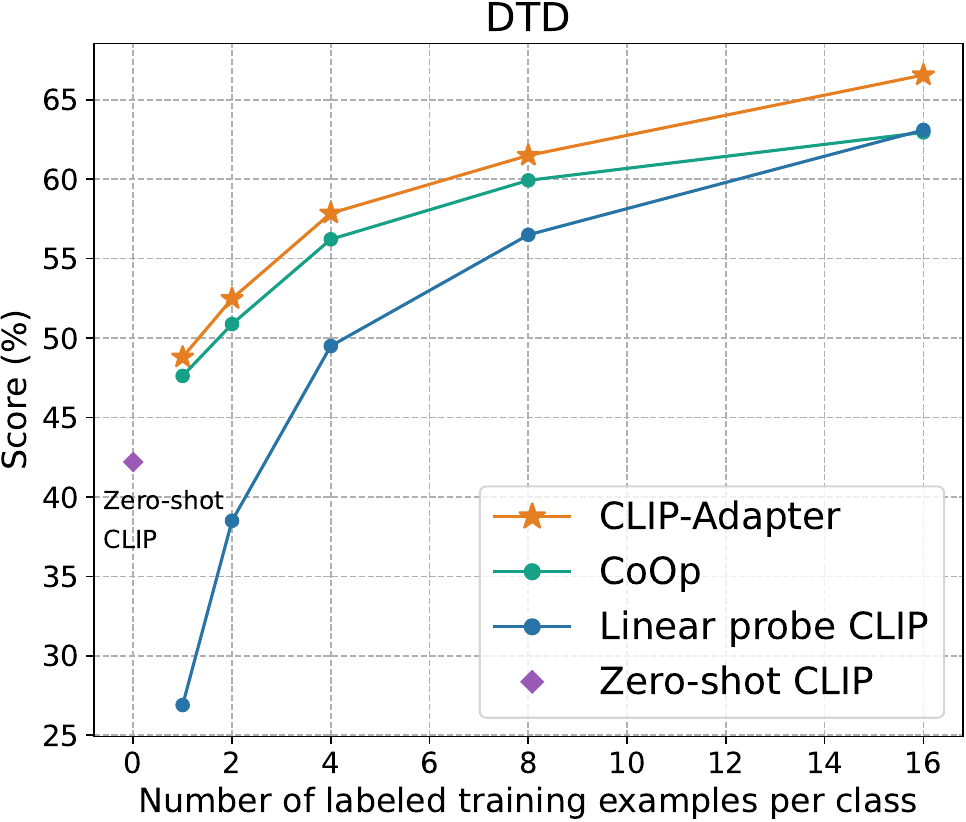}
    \end{minipage}
    \begin{minipage}[t]{0.325\linewidth}
    \centering
    \includegraphics[width=1.57in]{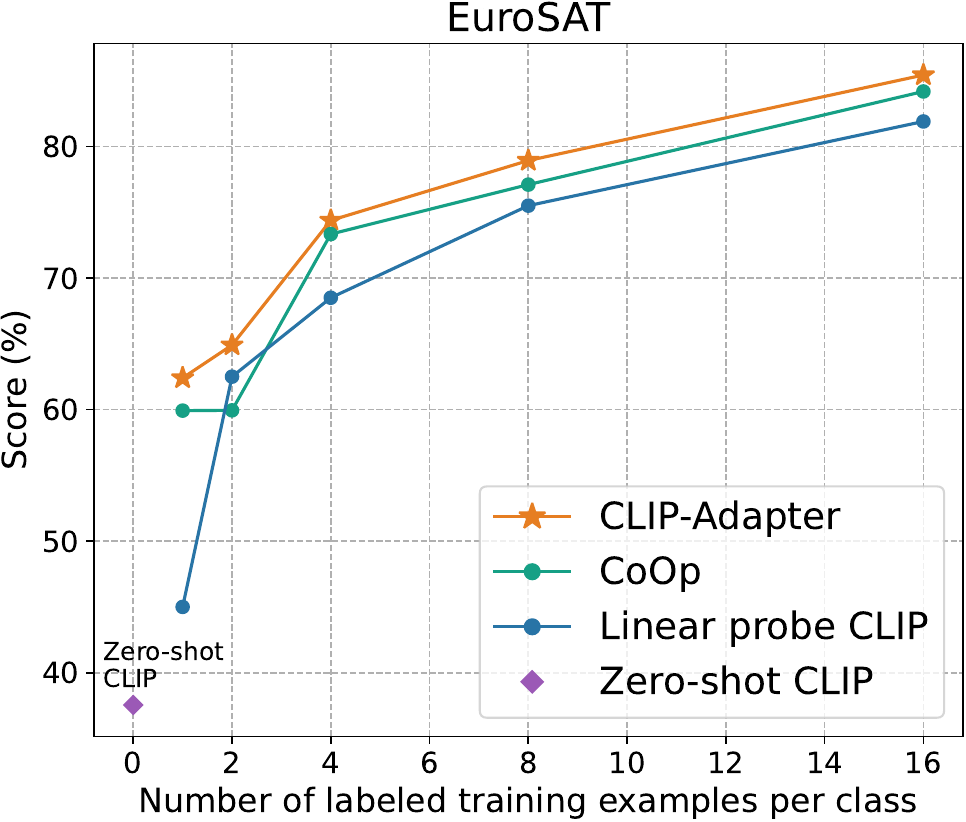}
    \end{minipage}
    \begin{minipage}[t]{0.325\linewidth}
    \centering
    \includegraphics[width=1.57in]{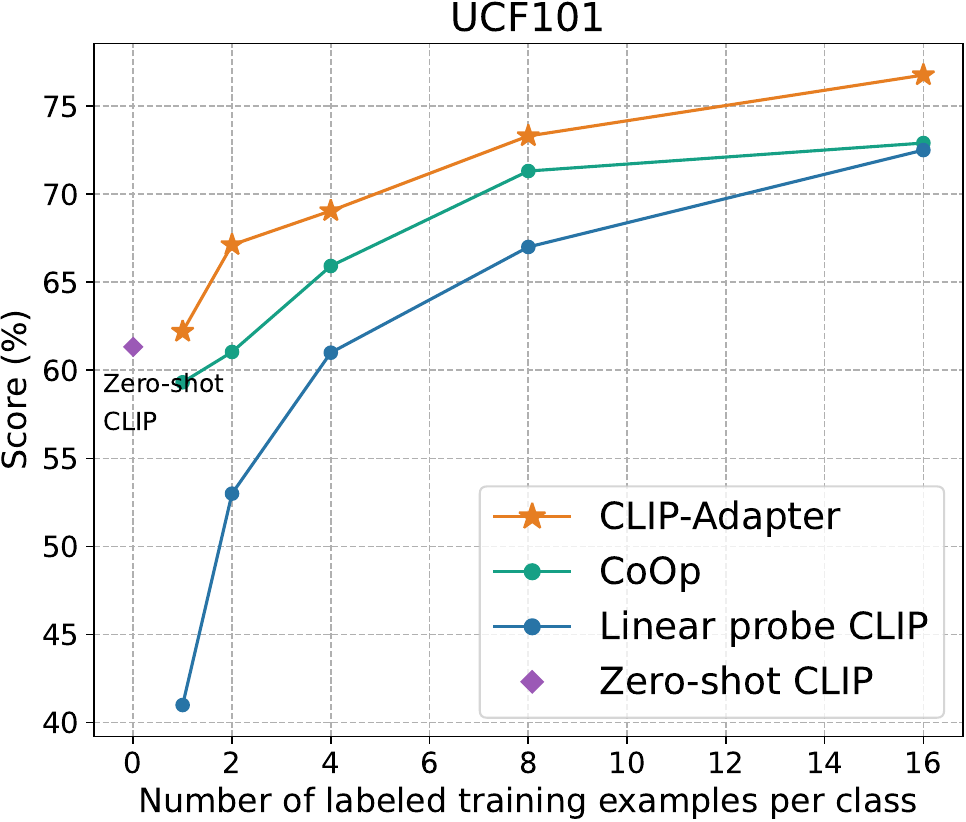}
    \end{minipage}
    \centering
    \vspace*{5pt}
    \caption{Results under CLIP-style preprocessing of few-shot learning on 11 datasets. Compared to CoOp-style preprocessing in Figure 3 of the main body, all results are boosted and CLIP-Adapter still shows leading performance over previous baselines across different training shots.}
    \label{fig:main_results_clip_preprocessing}
\end{figure*}

\section{Conclusions and Future Work}
We present CLIP-Adapter as an alternative of prompt-based approaches for few-shot image classification. The CLIP-Adapter revives the ``pretrain-finetuning'' paradigm by only fine-tuning a small number of additional bottleneck layers. To further improve the generalization ability, we adopt residual connections parameterized by a residual ratio to dynamically blend zero-shot knowledge with new adapted features. According to the experimental results, CLIP-Adapter outperforms competitive baselines on eleven image classification datasets under different few-shot setups. Extensive ablation studies confirm our design and prove CLIP-Adapter's ability in learning better feature manifolds. In the future, we plan to extend CLIP-Adapter to more vision-language applications and tasks.~We will also combine CLIP-Adapter with cache models~\cite{zhang2022tip,zhu2023not} and enhanced prompts~\cite{zhang2023prompt} to further unleash the power of CLIP backbone. 

\section*{Data Availability Statement}
No new data were created during the study. All experiments of this manuscript were conducted (training and evaluation) on 11 publicly available image classification datasets~\cite{deng2009imagenet, krause20133d, soomro2012ucf101, fei2004learning, nilsback2008automated, xiao2010sun, cimpoi2014describing, helber2019eurosat, maji2013fine, parkhi2012cats, bossard2014food}.

\section*{Acknowledgement}
This project is funded in part by the National Natural Science Foundation of China (No.62206272), by the National Key R\&D Program of China Project (No.2022ZD0161100), by the Centre for Perceptual and Interactive Intelligence (CPII) Ltd under the Innovation and Technology Commission (ITC)’s InnoHK, and by the General Research Fund of Hong Kong RGC Project 14204021. Hongsheng Li is a PI of CPII under the InnoHK.

\clearpage
\appendix

\section{Appendix}
\label{sec:appendix-clip-preprocessing}
\paragraph{Result Comparison under CLIP-Style Preprocessing.}
In Figure~\ref{fig:main_results_clip_preprocessing}, we present the result comparison under CLIP-style preprocessing of few-shot learning on 11 datasets. Compared to CoOp-style preprocessing, the performances of all methods are improved under CLIP-style preprocessing. Similar to Figure 3 of the main body, CLIP-Adapter still outperforms other baselines across different shot settings.

\bibliographystyle{sn-basic}
\bibliography{custom}


\end{document}